\documentclass{article} 
\usepackage{iclr2027_conference,times}

\usepackage{amsmath,amsfonts,bm}

\def\eqref#1{equation~\ref{#1}}

\def\1{\bm{1}}

\DeclareMathAlphabet{\mathsfit}{\encodingdefault}{\sfdefault}{m}{sl}
\SetMathAlphabet{\mathsfit}{bold}{\encodingdefault}{\sfdefault}{bx}{n}

\usepackage{subcaption}
\usepackage{hyperref}
\usepackage{url}
\usepackage{booktabs}
\usepackage{graphicx}
\usepackage{multirow}
\usepackage[table]{xcolor}
\usepackage{booktabs}
\usepackage{rotating} 
\usepackage{adjustbox}
\usepackage{pdflscape}

\definecolor{datasetblue}{RGB}{221,235,247}
\definecolor{scenarioorange}{RGB}{252,228,214}
\definecolor{oursblue}{RGB}{235,243,255}
\definecolor{headergray}{RGB}{242,242,242}
\usepackage{tikz}
\usepackage{graphicx}
\usepackage{subcaption}
\usepackage{newtxtext,newtxmath}
\usepackage{xcolor}
\usepackage{longtable}

\definecolor{headerblue}{HTML}{244564}
\definecolor{groupblue}{HTML}{D9EAF4}
\definecolor{rowgray}{HTML}{F4F7F9}
\definecolor{bestblue}{HTML}{005A9C}
\newcommand{\best}[1]{\textcolor{bestblue}{\bfseries #1}}
\renewcommand{\arraystretch}{1.12}
\title{ReAl: Refine neural operators via aligning dynamical flow}

\author{
Jing Xu$^{1,\dagger}$ \quad
Zenglin Xu$^{1,2}$ \quad
Li Yi$^{1}$ \quad
Ruoxi Jiang$^{1,2,\dagger,*}$ \quad
Yuan Qi$^{1,2,*}$ \\
$^{1}$ Shanghai Academy of AI for Science , ~~~$^{2}$ Fudan University \\
$\dagger$ Project leader , * Corresponding author \\
\texttt{\{roxie\_jiang, qiyuan\}@fudan.edu.cn} \\
}

\iclrfinalcopy
\begin{document}

\maketitle

\pagestyle{plain}
\thispagestyle{plain}

\pagestyle{fancy}
\fancyhead{}
\thispagestyle{fancy}

\begin{abstract}
Neural partial differential equation (PDE) surrogates trained with mean squared error (MSE) can capture dominant dynamics but often lose high-frequency information and fine-scale structures. Generative models (e.g., Flow Matching) can capture the distribution of fine-scale details, yet stochastic errors may accumulate during autoregressive rollouts and compromise long-term stability. 
To address these limitations, we introduce \textbf{Re}fine and \textbf{Al}ign
(\textbf{ReAl}), a model-agnostic plug-in that augments 
pretrained PDE
surrogates to improve the reconstruction of high-frequency structures
associated with complex physical phenomena.
In the refinement stage, a conditional flow-matching model
learns the residual distribution around the deterministic base
prediction, recovering missing fine-scale structures while retaining the dominant dynamics. 
In the alignment stage, to mitigate the generative discrepancies amplified in the autoregressive rollout, particularly in chaotic systems, we generate multiple trajectories from the same input and reinforce the refiner with trajectory-level rewards, thereby achieving long-horizon dynamical alignment.
Experiments across diverse surrogate architectures and
PDE tasks demonstrate that \textbf{ReAl} consistently improves high-frequency
fidelity, forecasting accuracy, and long-horizon stability.

\end{abstract}



\section{Introduction}



Partial differential equations (PDEs) are fundamental to modeling physical systems across fluid mechanics~\citep{davidson2004turbulence05}, heat and mass transfer~\citep{bergman2011fundamentals}, electromagnetics~\citep{schwinger1998classical07}, and chemical diffusion~\citep{crank1979mathematics08}. Classical finite-difference~\citep{morton1994numerical01} and finite-element methods~\citep{hughes2003finite02} provide reliable solutions, but their computational costs grow rapidly with spatial resolution, dimensionality, and parameter-space size. This limitation has driven the development of deep-learning-based PDE surrogates and neural operators, including DeepONet~\citep{lu2021learning03}, the Fourier Neural Operator~\citep{li2020fourier04}, and the Physics-Informed Neural Operator~\citep{goswami2023physics09}. By approximating PDE solution operators, these methods enable substantially faster inference for large-scale simulation and real-time scientific prediction.

Most deep-learning-based PDE surrogates are trained with mean squared error (MSE) objective, which provides stable optimization and effectively captures large-scale, low-frequency dynamics. 
However, as a mean-regression objective, MSE tends to suppress
high-frequency information, local vortices, and sharp interfaces
~\citep{lippe2023pde11,fotiadis2024stochastic12}, particularly on real-world datasets such as RealPDEBench~\citep{hu2026realpdebench}, where measurement noise and unresolved physical processes further
complicate the underlying dynamics.
Generative models~\citep{ho2020denoising19, rombach2022high20, lipman2022flow21} can alleviate this limitation by modeling the
conditional distribution of PDE solutions instead of producing only
one deterministic prediction, enabling richer high-frequency details~\citep{yao2026guided14,utkarsh2026physics,hu2025wavelet16}. However, stochastic sampling may introduce predictive variance and compromise the dynamical consistency and long-term stability required by deterministic PDE systems.

Motivated by these observations, we adopt a complementary modeling strategy: retain the MSE-trained surrogate as a stable predictor of dominant dynamics and apply a conditional flow-based refiner to its output. Rather than regenerating the entire PDE state, the refiner is
conditioned on the base prediction and learns to generate its
fine-scale residual correction~\citep{fotiadis2024stochastic12}. As shown in Fig.\ref{fig:intro}~\subref{fig:intro1}, this strategy recovers high-frequency information lost by MSE training while preserving low-frequency dynamics consistent in PDE predictions. Despite these benefits, a generative refiner can only partially
mitigate the rollout shift rather than completely eliminate it. Stochastic perturbations can accumulate during autoregressive rollout, creating a gap between recovering fine-scale structures at individual steps and maintaining forecasting accuracy over the resulting trajectory~\citep{luo2025pdespectralrefiner, Yoo2026DiffusionRolloutUR13}. As shown in Fig.\ref{fig:intro}~\subref{fig:intro2}, the base model preserves low-frequency dynamics over longer horizons, while high-frequency errors progressively accumulate and amplify, degrading long-term stability and physical consistency.

These observations motivate optimizing the refiner beyond individual
prediction quality by accounting for its cumulative effect on the
entire autoregressive trajectory. To this end, inspired by reinforcement learning for diffusion models~\citep{Liu2025FlowGRPOTF17,zheng2026diffusionnft18},
we introduce trajectory-alignment post-training for the flow-based
refiner. Starting from the same input state, we sample multiple
autoregressive trajectories using independent random seeds and evaluate their long-term quality with trajectory-level rewards. These rewards guide flow-matching updates to the same refiner, favoring refinements that yield accurate long-horizon forecasts. This extends the optimization target from
one-step accuracy to long-term trajectory quality without sacrificing
the refiner's ability to recover fine-scale structures. Our main
contributions are summarized as follows:

\begin{figure*}[t]
    \centering
    \begin{subfigure}[t]{0.435\textwidth}
        \centering
        \includegraphics[width=\linewidth]{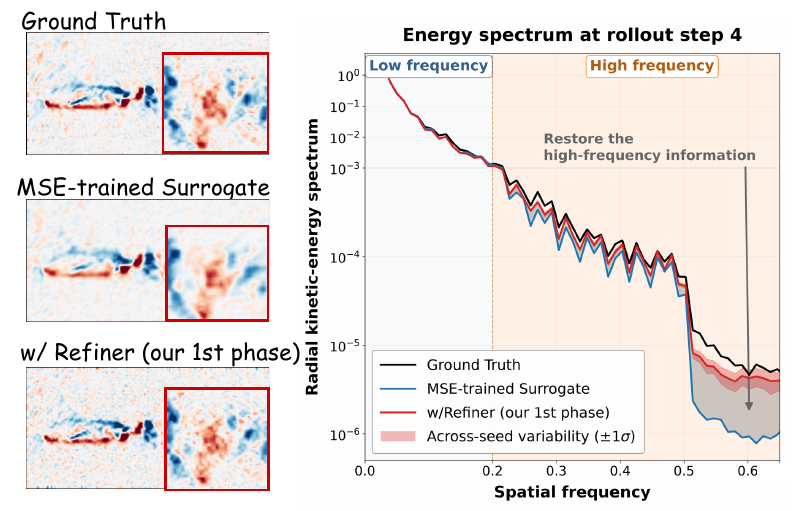}
        \caption{Necessity of refinement}
        \label{fig:intro1}
    \end{subfigure}
    \hfill
    \begin{subfigure}[t]{0.535\textwidth}
        \centering
        \includegraphics[width=\linewidth]{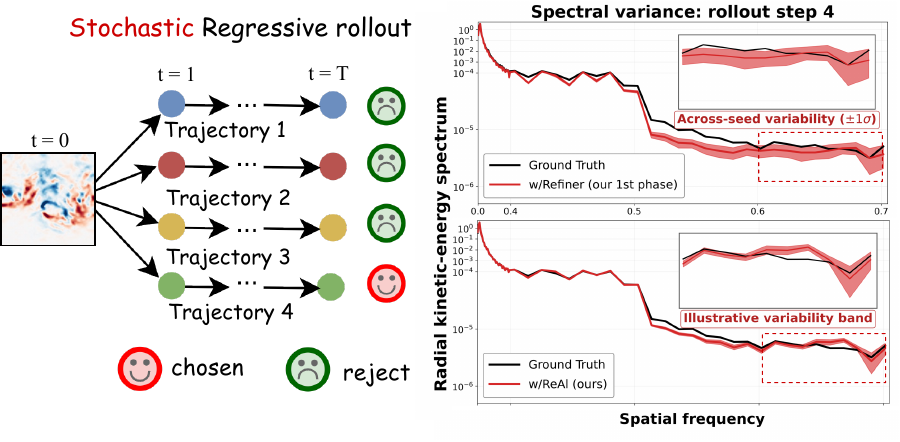}
        \caption{Necessity of alignment}
        \label{fig:intro2}
    \end{subfigure}
    \caption{(a) MSE training leads to high-frequency information loss, while stochastic refiner recovers the structures. (b) Sampling variability accumulates during stochastic autoregressive rollout, especially in the high-frequency regime. Our \textbf{ReAl} mitigates this accumulation of sampling variability.}
    \label{fig:intro}
\end{figure*}


\begin{itemize}
\item We propose \textbf{Re}fine and \textbf{Al}ign \textbf{(ReAl)}, a lightweight, model-agnostic PDE forecasting plug-in based on Flow Matching. \textbf{ReAl} addresses the loss of high-frequency information and fine-scale structures commonly observed in MSE-pretrained surrogates on real-world data, refining their predictions without modifying the underlying architectures.

\item To the best of our knowledge, we are the first to post-train long-term autoregressive PDE trajectories from a reinforcement learning perspective. \textbf{ReAl} samples multiple stochastic rollouts from the same initial state and optimizes the Flow refiner with trajectory-level rewards, aligning local residual corrections with long-term dynamical evolution.

\item Extensive experiments demonstrate that \textbf{ReAl} improves both prediction accuracy and long-term stability. Compared with the base models, it reduces the average RMSE by $\mathbf{34.3\%}$ at the $1$-st rollout step and by $\mathbf{45.4\%}$ at the $20$-th step, with larger gains over longer horizons on RealPDEBench. \textbf{ReAl} also recovers high-frequency physical structures, suppresses long-term error accumulation, and mitigates numerical dissipation.
\end{itemize}


\section{Related Works}

\textbf{Neural PDE Surrogates.}
Neural PDE surrogates approximate expensive numerical solvers by learning mappings between physical fields. Representative approaches include DeepONet~\citep{lu2021learning03}, the Fourier Neural Operator (FNO)~\citep{li2020fourier}, and physics-informed neural operators~\citep{goswami2023physics09}. Subsequent architectures have explored latent spectral representations, geometry-aware attention, hierarchical modeling, and large-scale pretraining, including LSM~\citep{wu2023solving}, DPOT~\citep{pmlr-v235-hao24d}, Transolver~\citep{wu2024transolver}, Poseidon~\citep{herde2024poseidon}, and HINE~\citep{jiang2026hierarchical}. Although these models are effective for autoregressive forecasting, prediction errors and insufficiently modeled frequency components can accumulate over long rollouts~\citep{lippe2023pde11,Yoo2026DiffusionRolloutUR13}. We retain such models as deterministic backbones and improve their predictions through a model-agnostic refinement plug-in.

\textbf{Generative Models and Refinements for PDEs.}
Diffusion and Flow Matching models
~\citep{ho2020denoising19,rombach2022high20,lipman2022flow21} have been widely applied to PDE simulation, data assimilation, and
physics-constrained generation ~\citep{shysheya2024conditional10,yao2026guided14,utkarsh2026physics}. DiffusionPDE addresses PDE solving under partial observations
~\citep{huang2024diffusionpde}, DYffusion targets probabilistic
multi-step forecasting~\citep{ruhling2023dyffusion}, and autoregressive
conditional diffusion models have been studied for turbulent-flow
prediction~\citep{kohl2026benchmarking}. Other methods improve multiscale reconstruction through wavelet or latent representations~\citep{hu2025wavelet16,Li2026DiBiLPSDI15}.
PDE-Refiner~\citep{lippe2023pde11}, Adaptive Flow Matching
~\citep{fotiadis2024stochastic12}, PDE-SpectralRefiner
~\citep{luo2025pdespectralrefiner}, and FlowRefiner
~\citep{dai2026flowrefiner} further recover non-dominant frequency
components through generative refinement. 
Unlike those methods based primarily on short-horizon objectives, ReAl directly optimizes long-term autoregressive rollouts.

\textbf{Post-Training of Diffusion Models.}
Reward-based post-training aligns diffusion and Flow Matching models with task-specific objectives. DDPO~\citep{black2024training} and DPOK~\citep{fan2023dpok} optimize diffusion sampling through reinforcement learning, Flow-GRPO~\citep{Liu2025FlowGRPOTF17} compares groups of samples, and DiffusionNFT~\citep{zheng2026diffusionnft18} employs negative-aware online reinforcement learning. We extend these ideas to PDE forecasting by post-training the refiner with trajectory-level rewards for long-term accuracy, stability, and physical consistency.

\section{Method}
\label{sec:method}

\begin{figure*}[t]
    \centering
    \includegraphics[width=0.95\textwidth,
        trim=0.5cm 0cm 0cm 0cm,
        clip]{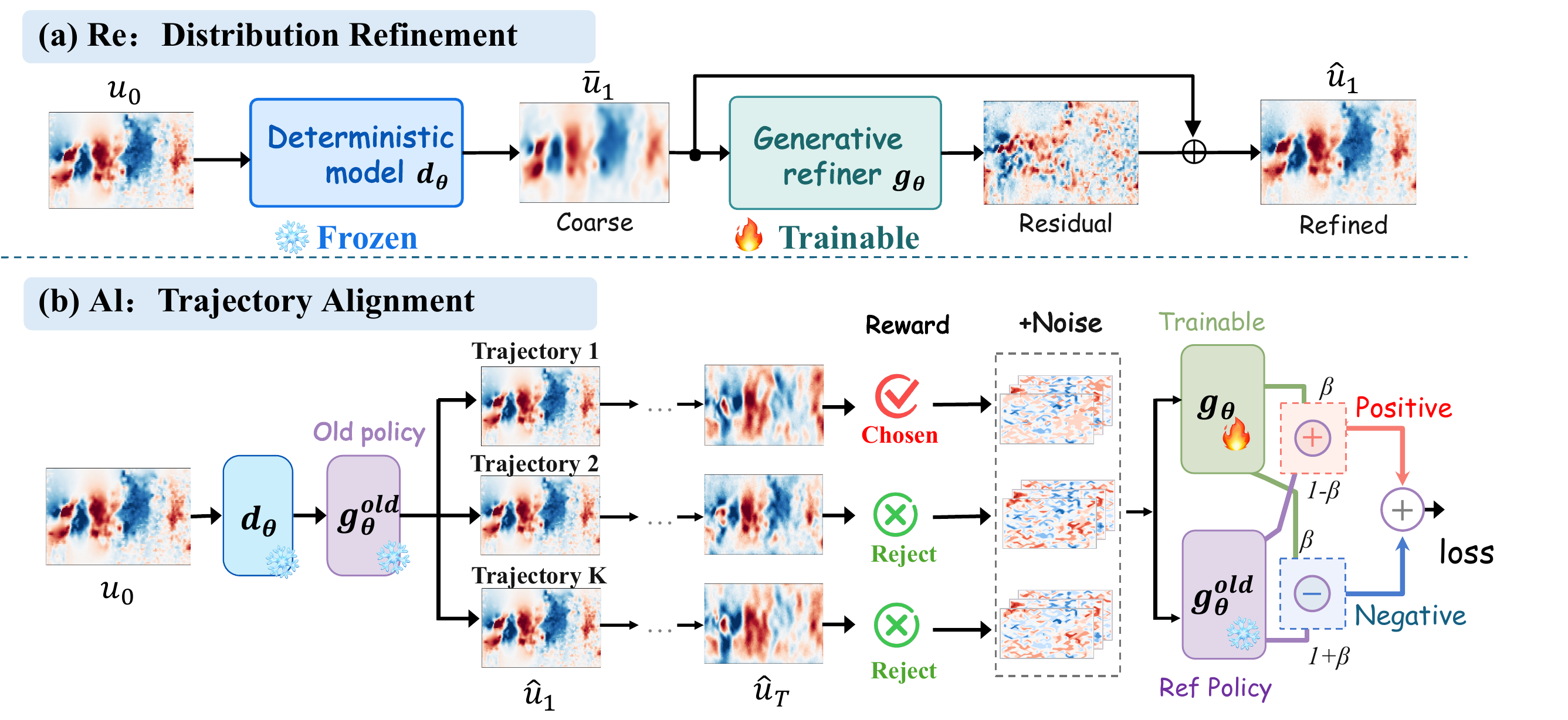}
    \caption{Overview of \textbf{Re}fine and \textbf{Al}ign (\textbf{ReAl}), a model-agnostic plug-in
for pretrained PDE surrogates. (a) \textbf{Refine}: a conditional flow matching 
refiner generates fine-scale residuals for the frozen deterministic
prediction. (b) \textbf{Align}: trajectory rewards from multiple
stochastic rollouts post-train the refiner for stable long-term PDE
forecasting. Both stages optimize the same conditional velocity field, with trajectory rewards weighting its flow-matching objectives during alignment.}
    \label{fig:trajectory_nft1}
\end{figure*}

In this section, we introduce \textbf{Re}fine and \textbf{Al}ign (\textbf{ReAl}), a model-agnostic plug-in for stable and accurate long-term partial differential equation forecasting, as illustrated in Fig.~\ref{fig:trajectory_nft1}. During refinement, we pretrain a flow matching refiner to correct the residual distribution of the base predictions. During alignment, we employ trajectory-level post-training to improve long-term forecasting stability.

\subsection{Preliminaries}
\noindent\textbf{PDE surrogates for forecasting.}
Consider a nonlinear dynamical system governed by a partial differential equation (PDE):
\begin{equation}
\frac{\partial \boldsymbol{u}}{\partial t}
=
\mathcal{G}(\boldsymbol{u};\phi),
\label{eq1:pde_dynamics}
\end{equation}
where \(\boldsymbol{u}(\boldsymbol{x},t)\) denotes a continuous physical field depending on spatial coordinate \(\boldsymbol{x}\) and time $t$, \(\phi\) represents the physical parameters governing the dynamics of the system, and $
\mathcal{G}: \mathcal{U} \times \Phi \rightarrow\mathcal{U}$ is a differential operator. By applying numerical discretization techniques, such as finite-difference or spectral methods, we obtain a discrete trajectory of consecutive observed states:$\{\boldsymbol{u}_0,\boldsymbol{u}_1,\ldots,\boldsymbol{u}_N\}$. Our goal is to learn a neural surrogate
model $f_\theta$, parameterized by $\theta$, that approximates the discrete
evolution operator and predicts the next state
$\boldsymbol{u}_{n+1}$ from the current state
$\boldsymbol{u}_n $\footnote{To simplify the notation, we consider one-step forecasting as a
representative setting. Our method is agnostic to the temporal
input--output configuration and can be applied to surrogates that use
multiple historical states or predict multiple future states.},
$\hat{\boldsymbol{u}}_{n+1}=f_\theta(\boldsymbol{u}_n).$
 For multi-step forecasting, the surrogate model is applied recursively in an autoregressive manner. The prediction at the \(k\)-th future step is given by $\hat{\boldsymbol{u}}_{n+k}
= f_{\theta}^{\circ k}(\boldsymbol{u}_n),$ where $f_{\theta}^{ \circ k}(\cdot)$ denotes the $k$-fold composition of
$f_{\theta}(\cdot)$.


\noindent\textbf{Conditional Flow Matching.} 
Conditional flow matching (CFM)~\citep{lipman2022flow21} learns a conditional vector field that
transports samples from a Gaussian noise distribution to the target
data distribution conditioned on $\boldsymbol{c}$. During training, an
intermediate state is sampled along a linear interpolation between a
noise sample and a target sample, and the vector field
$\boldsymbol{v}_{\psi}$ is optimized to predict the corresponding
transport velocity. At inference, the learned vector field is integrated
from noise to data to generate a sample under the given condition.
Here, the flow time $\tau\in[0,1]$ parameterizes the generative transport
process and is distinct from the physical time of PDE evolution.



\subsection{Distribution Refinement with Conditional Flow Matching}
\label{sec:stochastic_refinement}

We first introduce the distribution refinement phase of \textbf{ReAl}, as shown in Fig.~\ref{fig:trajectory_nft1}(a). Most existing PDE surrogates are trained with MSE, enabling the deterministic predictor $\boldsymbol{d}_{\theta}$ to stably capture dominant dynamics but often suppressing high-frequency structures through mean regression. To address this limitation, we introduce a conditional flow matching refiner $\boldsymbol{g}_{\psi}$ that learns the residual distribution between the base prediction and ground truth, recovering fine-scale structures.


\noindent\textbf{Deterministic Base Prediction.}
Given the current PDE state $\boldsymbol{u}_n$, the MSE-based
deterministic predictor produces the next-state estimate, $\bar{\boldsymbol{u}}_n$, and is
pretrained by minimizing its discrepancy from the ground truth, $\boldsymbol{u}_{n+1}$:

\begin{equation}
\begin{aligned}
\bar{\boldsymbol{u}}_{n+1}=
\boldsymbol{d}_{\theta}(\boldsymbol{u}_n),~~~~
\mathcal{L}_{\mathrm{MSE}}(\theta)
=
\mathbb{E}
\left[
\left\|
\bar{\boldsymbol{u}}_{n+1}
-
\boldsymbol{u}_{n+1}
\right\|^2
\right].
\end{aligned}
\label{eq:deterministic_predictor}
\end{equation}

The optimal MSE predictor approaches the conditional mean,
   $
    \mathbb{E}
    \left[
        \boldsymbol{u}_{n+1}\mid\boldsymbol{u}_n
    \right].$ Although this reliably captures the dominant dynamics, conditional averaging tends to suppress non-dominant fine-scale structures, resulting in overly smooth predictions.

\noindent\textbf{Conditional Flow Refinement.}
To recover non-dominant structures suppressed by MSE training, we introduce a conditional flow matching refiner \(\boldsymbol{g}_{\psi}\) that models the conditional distribution of fine-scale residuals. Specifically, we define the target residual, $\boldsymbol{r}_{n+1}$, and the condition, $\boldsymbol{c}_n$, as:
\begin{equation}
\begin{aligned}
\boldsymbol{r}_{n+1}
=
\boldsymbol{u}_{n+1}
-
\bar{\boldsymbol{u}}_{n+1},~~~~
\boldsymbol{c}_n
=
\operatorname{Concat}
\left[
\boldsymbol{u}_n,
\bar{\boldsymbol{u}}_{n+1}
\right].
\end{aligned}
\end{equation}

The refiner learns to generate \(\boldsymbol{r}_{n+1}\) conditioned on the current state, $\boldsymbol{u}_{n}$, and deterministic base prediction, $\bar{\boldsymbol{u}}_{n+1}$.
Given a Gaussian sample \(\boldsymbol{z}_0\sim\mathcal{N}(\boldsymbol{0},\boldsymbol{I})\), we construct a linear flow path toward the target residual:
\begin{equation}
\boldsymbol{z}_{\tau}
=
(1-\tau)\boldsymbol{z}_0
+
\tau\boldsymbol{r}_{n+1},
\qquad
\tau\sim\mathcal{U}(0,1).
\label{eq:method_z}
\end{equation}

The conditional velocity field \(\boldsymbol{v}_{\psi}\) is trained to match the corresponding transport velocity:
\begin{equation}
\mathcal{L}_{\mathrm{refine}}(\psi)
=
\mathbb{E}
\left[
\left\|
\boldsymbol{v}_{\psi}
\left(
\boldsymbol{z}_{\tau},
\tau
\mid
\boldsymbol{c}_n
\right)
-
\left(
\boldsymbol{r}_{n+1}
-
\boldsymbol{z}_0
\right)
\right\|^2
\right].
\end{equation}

At inference, the generative mapping \(\boldsymbol{g}_{\psi}\) is obtained by integrating \(\boldsymbol{v}_{\psi}\) from \(\tau=0\) to \(\tau=1\), starting from \(\boldsymbol{\epsilon}\sim\mathcal{N}(\boldsymbol{0},\boldsymbol{I})\). The generated residual and refined prediction are:
\begin{equation}
\begin{aligned}
\hat{\boldsymbol{r}}_{n+1}
=
\boldsymbol{g}_{\psi}
\left(
\boldsymbol{\epsilon}
\mid
\boldsymbol{c}_n
\right),~~~~
\hat{\boldsymbol{u}}_{n+1}
=
\bar{\boldsymbol{u}}_{n+1}
+
\hat{\boldsymbol{r}}_{n+1}.
\end{aligned}
\end{equation}
 The deterministic predictor and stochastic refiner are complementary: the former captures stable dominant dynamics, while the latter restores residual fine-scale structures without imposing a strict low- and high-frequency decomposition, as illustrated in Fig.~\ref{fig:trajectory_nft1}(a).

\begin{figure*}[t]
    \centering
    \includegraphics[width=\textwidth]{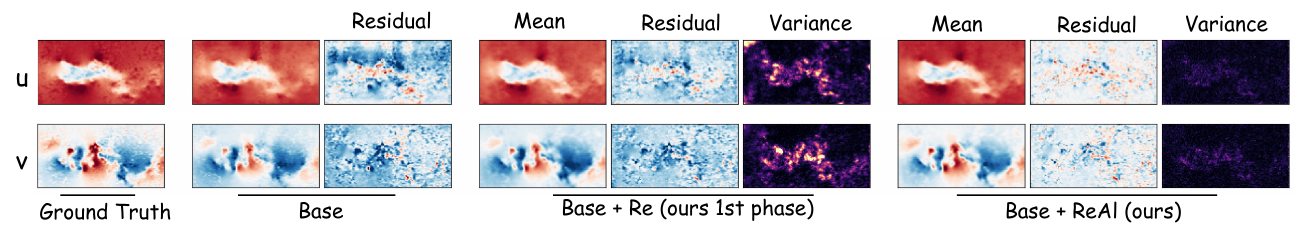}
    \caption{Qualitative results on Controlled Cylinder \textit{Real2Real} for velocity components \(u\) and \(v\). Residuals are computed against the ground truth, while variance is measured across random seeds. Base+ReAl produces more accurate and consistent predictions.
    }
    \label{fig:method_mean_var}
\end{figure*}

\subsection{Trajectory Alignment for PDE Forecasting}

Although combining the deterministic predictor
$\boldsymbol{d}_{\theta}$ with a conditional flow matching refiner
$\boldsymbol{g}_{\psi}$ recovers non-dominant fine-scale information
missing from the base prediction, perturbations introduced by stochastic
flow sampling may accumulate during autoregressive rollouts and degrade
long-term stability. To address this issue, we introduce the
trajectory-alignment post-training phase of \textbf{ReAl}, as shown in Fig.~\ref{fig:trajectory_nft1}(b). Specifically,
we keep $\boldsymbol{d}_{\theta}$ frozen and post-train the refiner
$\boldsymbol{g}_{\psi}$ using trajectory-level rewards from multiple
stochastic rollouts, thereby improving the quality of long-term PDE
trajectories.

\noindent\textbf{Trajectory rollout.}
Given an initial state \(\boldsymbol{u}_n\), we generate \(K\) candidate trajectories using the frozen deterministic predictor \(\boldsymbol{d}_{\theta}\) and a snapshot of the refiner \(\boldsymbol{g}_{\psi}^{\mathrm{old}}\). All candidates start from the same state but use independent noise sequences. For the \(i\)-th candidate, we initialize \(\hat{\boldsymbol{u}}^{(i)}_n=\boldsymbol{u}_n\) and recursively apply the refined transition:
\begin{equation}
\begin{aligned}
\bar{\boldsymbol{u}}^{(i)}_{n+k}
&=
\boldsymbol{d}_{\theta}
\left(
\hat{\boldsymbol{u}}^{(i)}_{n+k-1}
\right),
\boldsymbol{c}^{(i)}_{n+k-1}
=
\operatorname{Concat}
\left[
\hat{\boldsymbol{u}}^{(i)}_{n+k-1},
\bar{\boldsymbol{u}}^{(i)}_{n+k}
\right], \\
\hat{\boldsymbol{r}}^{(i)}_{n+k}
&=
\boldsymbol{g}_{\psi^{\mathrm{old}}}
\left(
\boldsymbol{\epsilon}^{(i)}_k
\mid
\boldsymbol{c}^{(i)}_{n+k-1}
\right),
\quad
\boldsymbol{\epsilon}^{(i)}_k
\sim
\mathcal{N}(\boldsymbol{0},\boldsymbol{I}), \\
\hat{\boldsymbol{u}}^{(i)}_{n+k}
&=
\bar{\boldsymbol{u}}^{(i)}_{n+k}
+
\hat{\boldsymbol{r}}^{(i)}_{n+k},
\quad
k=1,\ldots,h.
\end{aligned}
\end{equation}

After \(h\) autoregressive steps, the resulting trajectory is:
\begin{equation}
\widehat{\mathcal{T}}^{(i)}_n
=
\left[
\hat{\boldsymbol{u}}^{(i)}_{n+1},
\hat{\boldsymbol{u}}^{(i)}_{n+2},
\ldots,
\hat{\boldsymbol{u}}^{(i)}_{n+h}
\right],
\qquad
i=1,\ldots,K.
\end{equation}
Despite sharing the same initial state and deterministic predictor, independent residual samples produce different long-term trajectories. As shown in Fig.~\ref{fig:method_mean_var}, stochastic refinement (\(\text{Base}+\text{Re}\)) introduces high-frequency variance that can accumulate during autoregressive rollouts.

\noindent\textbf{Trajectory-level reward.}
Each candidate trajectory $\widehat{\mathcal{T}}^{(i)}_n$ is evaluated against the corresponding ground-truth trajectory $\mathcal{T}_n$ to obtain its trajectory-level reward:

\begin{equation}
\begin{aligned}
R^{(i)}
=
\mathcal{R}
\left(
\widehat{\mathcal{T}}^{(i)}_n,
\mathcal{T}_n
\right), \quad
R^{(i)}_{\mathrm{MSE}}
=
-
\left\|
\hat{\boldsymbol{u}}^{(i)}_{n+h}
-
\boldsymbol{u}_{n+h}
\right\|^2,
\end{aligned}
\label{eq:trajectory_reward}
\end{equation}
where \(R^{(i)}_{\mathrm{MSE}}\) denotes the negative mean squared error at the terminal state, such that a larger reward indicates a more accurate terminal prediction. Although the reward is evaluated only at the endpoint, the terminal state reflects the accumulated effects of stochastic residual corrections propagated throughout the multi-step autoregressive rollout. It therefore serves as a measure of long-term trajectory quality. More generally, \(\mathcal{R}\) may be defined using discrepancies in physically relevant quantities, such as vorticity, divergence, and kinetic energy, or by combining multiple criteria to evaluate complementary properties of the predicted dynamics.

Since reward magnitudes can vary across initial conditions, we
standardize the rewards within each group of $K$ trajectories. Let $\mu_R$ and $\sigma_R$ denote the group
mean and standard deviation, respectively. We then clip the standardized
reward and linearly rescale it to $[0,1]$:
\begin{equation}
w^{(i)}
=
\frac{1}{2}
\left[
1+
\frac{1}{a_{\max}}
\operatorname{clip}
\left(
\frac{R^{(i)}-\mu_R}
{\sigma_R+\epsilon_{\mathrm{num}}},
-a_{\max},
a_{\max}
\right)
\right].
\label{eq:reward_weight}
\end{equation}
Here, $\epsilon_{\mathrm{num}}$ is a small constant that ensures
numerical stability and \(a_{\max}>0\) controls the clipping range. Consequently, higher-quality trajectories receive
weights closer to $1$, whereas lower-quality receive
weights closer to $0$.

\noindent\textbf{Trajectory reward-Guided Flow Alignment.}
Trajectory alignment post-trains the conditional velocity field learned during refinement (Sec.~\ref{sec:stochastic_refinement}). We incorporate feedback from autoregressive rollouts through positive and negative flow-matching terms constructed from the current and old velocity fields, weighted by trajectory rewards. For the $k$-th autoregressive transition of the $i$-th candidate
trajectory, $\boldsymbol{z}^{(i,k)}_{\tau}$ denotes the intermediate
residual state at flow time $\tau$, following
Eq.~\eqref{eq:method_z}. The velocity fields predicted by the
current policy, old policy, and reference model are uniformly written as:
\begin{equation}
\boldsymbol{v}^{s,(i,k)}
=
\boldsymbol{v}_{\psi^{s}}
\left(
\boldsymbol{z}^{(i,k)}_{\tau},
\tau
\mid
\boldsymbol{c}^{(i)}_{n+k-1}
\right),
\qquad
s\in
\left\{
\mathrm{cur},
\mathrm{old},
\mathrm{ref}
\right\}.
\label{eq:trajectory_velocity_fields}
\end{equation}
Here, $\psi^{\mathrm{cur}}\equiv\psi$ denotes the parameters of the
refiner being updated, $\psi^{\mathrm{old}}$ denotes the old-policy
parameters used to generate the candidate trajectories, and
$\psi^{\mathrm{ref}}$ denotes the parameters of the pretrained reference
model, which remain frozen throughout post-training. Following the negative-aware forward-process formulation of
DiffusionNFT~\citep{zheng2026diffusionnft18}, we construct positive and
negative velocity fields from the current- and old-policy predictions:
\begin{equation}
\begin{aligned}
\boldsymbol{v}^{+,(i,k)}
=
\beta\boldsymbol{v}^{\mathrm{cur},(i,k)}
+
(1-\beta)\boldsymbol{v}^{\mathrm{old},(i,k)}, \quad
\boldsymbol{v}^{-,(i,k)}
=
(1+\beta)\boldsymbol{v}^{\mathrm{old},(i,k)}
-
\beta\boldsymbol{v}^{\mathrm{cur},(i,k)},
\end{aligned}
\label{eq:positive_negative_velocity}
\end{equation}
where $\beta>0$ controls the deviation of the positive and negative
velocity fields from the old policy. The corresponding positive and negative flow matching losses are:
\begin{equation}
\ell_{\pm}^{(i,k)}
=
\left\|
\boldsymbol{v}^{\pm,(i,k)}
-
\left(
\boldsymbol{r}^{(i)}_{n+k}
-
\boldsymbol{z}^{(i,k)}_0
\right)
\right\|^2.
\label{eq:local_positive_negative_loss}
\end{equation}

The final alignment objective averages over both candidate trajectories
$i$ and autoregressive steps $k$:
\begin{equation}
\mathcal{L}_{\mathrm{align}}
=
\frac{1}{\beta}
\mathbb{E}_{i,k}
\left[
w^{(i)}\ell_{+}^{(i,k)}
+
\left(1-w^{(i)}\right)\ell_{-}^{(i,k)}
\right].
\label{eq:flow_alignment}
\end{equation}
Same trajectory-level reward \(w^{(i)}\) is applied to every autoregressive step of the \(i\)-th candidate. A high trajectory reward produces a weight close to \(1\), emphasizing the positive objective and encouraging the associated residual corrections. A low reward produces a weight closer to \(0\), emphasizing the negative objective. Because the same trajectory-level weight is shared at every rollout step, the terminal feedback supervises residual generation along the entire trajectory.

To stabilize post-training, we further regularize the refiner
toward the frozen pretrained reference model while retaining the
supervised residual-learning objective. The training objective is:
\begin{equation}
\mathcal{L}
=
\mathcal{L}_{\mathrm{align}}
+
\lambda_{\mathrm{ref}}
\mathbb{E}_{i,k}
\left[
\left\|
\boldsymbol{v}^{\mathrm{cur},(i,k)}
-
\boldsymbol{v}^{\mathrm{ref},(i,k)}
\right\|^2
\right]
+
\lambda_{\mathrm{sup}}
\mathbb{E}_{i,k}
\left[
\left\|
\hat{\boldsymbol{r}}^{(i)}_{n+k}
-
\boldsymbol{r}^{(i)}_{n+k}
\right\|^2
\right].
\label{eq:post_training_objective}
\end{equation}
The reference-model regularization prevents excessive deviation from the
pretrained flow dynamics, while the supervised term preserves the
refiner's residual-modeling capability. Throughout post-training, only
the current refiner parameters $\psi$ are updated; the deterministic
predictor $\boldsymbol{d}_{\theta}$, old-policy parameters
$\boldsymbol{g}_{\psi}^{\mathrm{old}}$, and reference-model parameters
$\boldsymbol{g}_{\psi}^{\mathrm{ref}}$ remain frozen.

Overall, \textbf{ReAl} produces more accurate long-term predictions with lower across-seed variance. By favoring residual corrections that remain reliable over long horizons, it produces more accurate predictions with lower across-seed variance, as shown in Fig.~\ref{fig:method_mean_var} (Base+ReAl).

\section{Experiment}

\textbf{Benchmarks.} 
We evaluate on RealPDEBench~\citep{hu2026realpdebench}, comprising four physical sub-datasets: Cylinder, Controlled Cylinder, Foil, and FSI, which provides paired numerical simulations and real-world measurements. Real observations contain noise, unresolved effects, and finer-scale dynamics absent from simulations. We consider two settings: \textit{Sim2Real}, training on simulations and testing on real data; and \textit{Real2Real}, training and testing on real data\footnote{For Cylinder, we exclude trajectory \texttt{3656.h5} due to a reproducible spatiotemporal misalignment in our processing pipeline.}.

\textbf{Evaluation metircs.}
To evaluate quality of prediction , we report the root mean squared error (RMSE) and relative (Rel. $L_2$) error. To further assess physical fidelity, we use the Fourier-space RMSE (fRMSE), frequency error (FE), and kinetic energy error (KE). Detailed definitions of these metrics are provided in the Appendix~\ref{sec:appendix_metric}.

\textbf{Experiment setup.}
In our experiments, following the task setting of RealPDEBench, we use 10 consecutive input frames to directly predict the subsequent 10 frames. We adopt U-Net~\citep{ronneberger2015u}, FNO~\citep{li2020fourier}, DPOT-S~\citep{pmlr-v235-hao24d}, and Transolver~\citep{wu2024transolver} as the base architectures. We then freeze the base model and train a lightweight time-conditioned U-Net flow matching Refiner (approximately 5.04M parameters, typically under one-tenth of the base model size) for 50,000 steps with a batch size of 16 and a learning rate of \(1\times10^{-4}\). Details are provided in Appendix~\ref{sec:appendix_architecture}. Subsequently, we initialize ReAl from the best Refiner checkpoint while keeping the base model frozen, and optimize ReAl for 10,000 steps using the trajectory-alignment objective. We use AdamW with a batch size of 4 and a learning rate of \(2\times10^{-5}\). For each condition, we sample \(K=4\) complete trajectories. Each trajectory consists of five autoregressive rollout blocks, with each block generated by integrating the flow velocity field using four Euler flow steps. We use the negative MSE of the final, fifth prediction block as the trajectory reward. We set \(\beta=0.1\) in Eq.~\ref{eq:flow_alignment}, and set \(\lambda_{\mathrm{ref}}=0.01\) and \(\lambda_{\mathrm{sup}}=0.1\) in Eq.~\ref{eq:post_training_objective}.


\subsection{Main Results}

\newcommand{\oursbetter}[2]{%
    \cellcolor{oursblue}%
    \begin{tikzpicture}[baseline=(value.base)]
        \node[
            minimum width=4.4em,
            minimum height=3.4ex,
            inner sep=0pt
        ] (box) {};

        \node[
            inner sep=0pt,
            font=\bfseries
        ] (value) at ([yshift=0.35ex]box.center) {#1};

        \node[
            anchor=south east,
            inner sep=0.2pt,
            font=\tiny,
            text=green!45!black
        ] at (box.south east) {+#2};
    \end{tikzpicture}%
}

\begin{table*}[t]
\centering
\caption{
Main experimental results on
RealPDEBench~\citep{hu2026realpdebench}.  We report RMSE over the
10 predicted future frames on test set. 
lower is better. The small percentage in the bottom-right
corner indicates the relative RMSE reduction over the corresponding
baseline.
}
\label{tab:main_results}

\setlength{\tabcolsep}{5.5pt}
\renewcommand{\arraystretch}{1.35}

\resizebox{\textwidth}{!}{
\begin{tabular}{lllcccccccc}
\toprule
\rowcolor{gray!15}
& & &
\multicolumn{2}{c}{\textbf{UNet}}
& \multicolumn{2}{c}{\textbf{FNO}}
& \multicolumn{2}{c}{\textbf{DPOT}}
& \multicolumn{2}{c}{\textbf{Transolver}}
\\

\rowcolor{gray!15}
\multirow{-2}{*}{\textbf{Benchmark}}
& \multirow{-2}{*}{\textbf{Setting}}
& \multirow{-2}{*}{\textbf{Task}}
& Baseline
& \cellcolor{oursblue}\textbf{+ Ours}
& Baseline
& \cellcolor{oursblue}\textbf{+ Ours}
& Baseline
& \cellcolor{oursblue}\textbf{+ Ours}
& Baseline
& \cellcolor{oursblue}\textbf{+ Ours}
\\
\midrule
\multirow{8}{*}{\textcolor{blue!60!black}{\textbf{RealPDEBench}}}
&
\multirow{4}{*}{\textcolor{orange!70!black}{\textit{Sim2Real}}}
& Cylinder
& 0.0256 & \oursbetter{0.0216}{16\%}
& 0.0425 & \oursbetter{0.0261}{39\%}
& 0.0235 & \oursbetter{0.0236}{1\%}
& 0.0315 & \oursbetter{0.0244}{23\%}
\\

& & Controlled Cylinder
& 0.0219 & \oursbetter{0.0155}{29\%}
& 0.0309 & \oursbetter{0.0168}{46\%}
& 0.0183 & \oursbetter{0.0144}{21\%}
& 0.0277 & \oursbetter{0.0147}{47\%}
\\

& & Foil
& 0.0167 & \oursbetter{0.0136}{19\%}
& 0.0197 & \oursbetter{0.0122}{38\%}
& 0.0290 & \oursbetter{0.0169}{42\%}
& 0.0345 & \oursbetter{0.0276}{20\%}
\\

& & FSI
& 0.0247 & \oursbetter{0.0161}{35\%}
& 0.0452 & \oursbetter{0.0132}{71\%}
& 0.0201 & \oursbetter{0.0116}{42\%}
& 0.0273 & \oursbetter{0.0128}{53\%}
\\

\cmidrule(lr){2-11}

&
\multirow{4}{*}{\textcolor{orange!70!black}{\textit{Real2Real}}}
& Cylinder
& 0.0082 & \oursbetter{0.0073}{11\%}
& 0.0135 & \oursbetter{0.0089}{34\%}
& 0.0115 & \oursbetter{0.0092}{20\%}
& 0.0154 & \oursbetter{0.0092}{41\%}
\\

& & Controlled Cylinder
& 0.0089 & \oursbetter{0.0073}{19\%}
& 0.0102 & \oursbetter{0.0086}{18\%}
& 0.0114 & \oursbetter{0.0081}{30\%}
& 0.0137 & \oursbetter{0.0083}{39\%}
\\

& & Foil
& 0.0082 & \oursbetter{0.0069}{14\%}
& 0.0134 & \oursbetter{0.0070}{48\%}
& 0.0144 & \oursbetter{0.0074}{49\%}
& 0.0241 & \oursbetter{0.0101}{58\%}
\\

& & FSI
& 0.0058 & \oursbetter{0.0053}{8\%}
& 0.0089 & \oursbetter{0.0056}{37\%}
& 0.0106 & \oursbetter{0.0056}{47\%}
& 0.0119 & \oursbetter{0.0056}{53\%}
\\

\bottomrule
\end{tabular}
}
\end{table*}

\begin{figure*}[t]
    \centering
    \includegraphics[width=\textwidth]{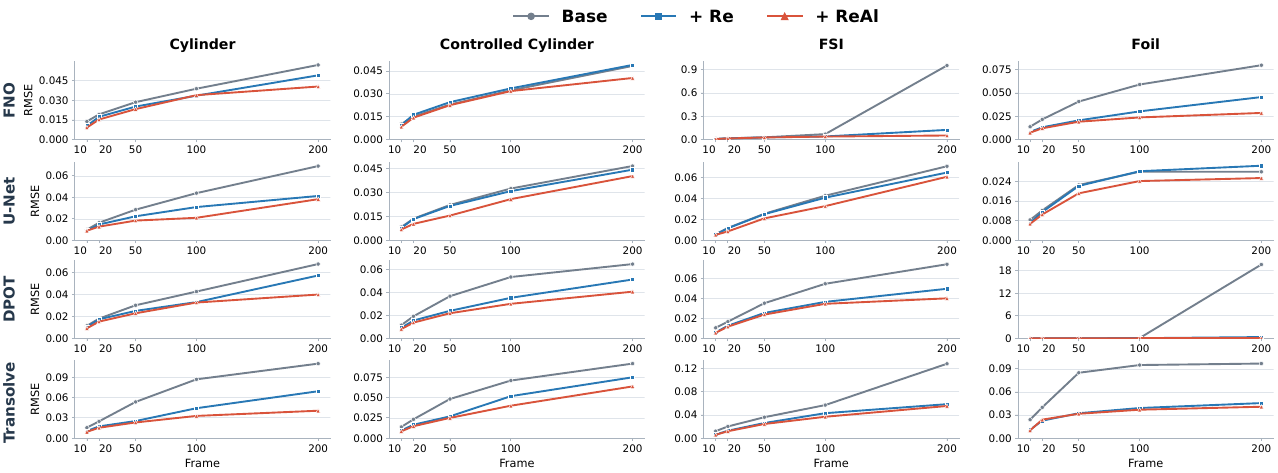}
    \caption{
        RMSE over long autoregressive rollout frames under the RealPDEBench \textit{Real2Real} setting. Long-horizon error accumulation causes severe degradation for several base
models, most notably DPOT on Foil and FNO on FSI. By contrast, Base+ReAl consistently
limits error growth and preserves stable predictions over long rollout horizons.
    }
    \label{fig:exp1}
\end{figure*}

\begin{figure*}[t]
    \centering
    \includegraphics[width=0.95\textwidth]{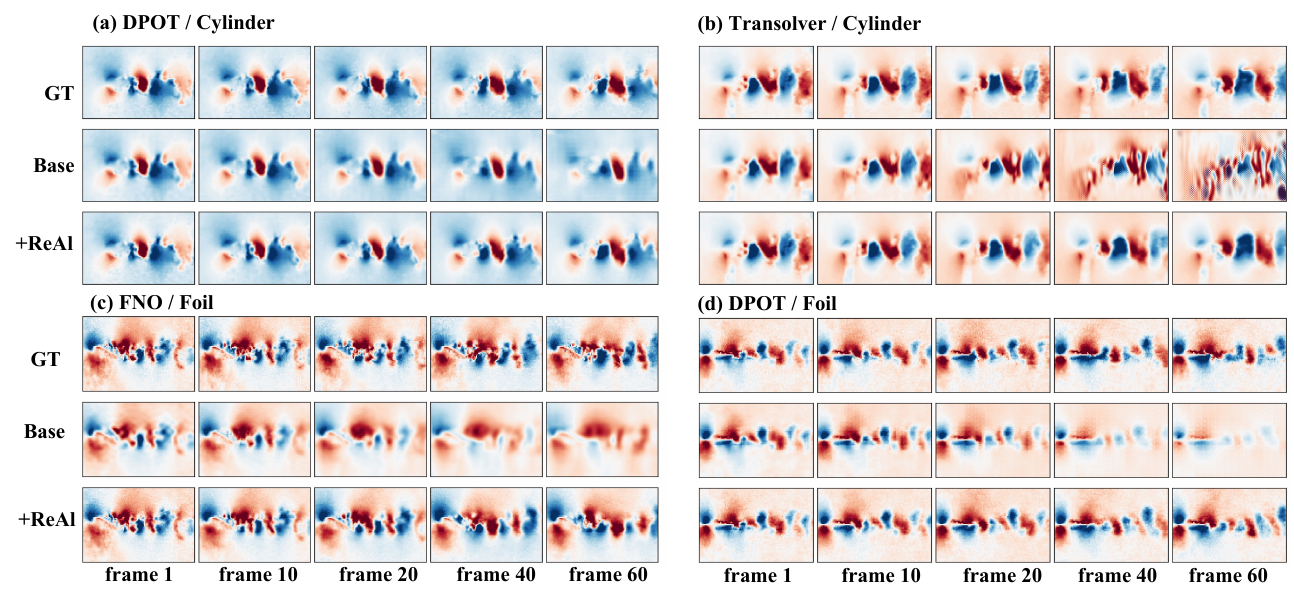}
    \caption{Visualization of long-horizon \textit{Real2Real} rollouts (v channel): Base+\textbf{ReAl} (a) restores fine-scale details, (b) suppresses error accumulation, and (c, d) mitigates numerical dissipation to preserve flow energy and wake structures.
    }
    \label{fig:expose_visualization}
\end{figure*}

\textbf{Prediction Accuracy.}
We first evaluate \textbf{ReAl} after a single rollout, i.e., direct prediction of one 10-frame future block, under the \textit{Sim2Real} and \textit{Real2Real} settings of RealPDEBench. As shown in Table~\ref{tab:main_results}, Base+ReAl consistently improves diverse base architectures across physical tasks. Averaged over all models, tasks, and settings, Base+ReAl reduces RMSE by $33.9\%$, including reductions of $33.7\%$ on \textit{Sim2Real} and $34.1\%$ on \textit{Real2Real}. Notably, ReAl achieves a $71\%$ relative improvement for FNO on the \textit{Sim2Real} FSI task. These results demonstrate that Base+ReAl provides effective post-training refinement even after a single rollout.


\textbf{Long-Term Stability.} To further evaluate long-term stability, we extend the autoregressive prediction horizon to 200 frames and compare the base models, Base$+$Re with only the flow refiner, and Base$+$ReAl with additional trajectory-alignment post-training in Fig.~\ref{fig:exp1}. As the rollout horizon increases, prediction errors of the base models accumulate continuously, with some models exhibiting severe error blow-up, such as FNO on FSI and DPOT on Foil. For DPOT on Foil, Base+ReAl limits 200-frame error growth to $10.9\times$, compared with $44.5\times$ for Base$+$Re and $1{,}360\times$ for the base model. Although Base$+$Re reduces the overall error through fine-scale residual correction, perturbations introduced by stochastic sampling may still be progressively amplified during long-term forecasting. In contrast, Base$+$ReAl exhibits substantially slower error growth across most models and tasks and effectively suppresses these instabilities. Fig.~\ref{fig:expose_visualization} compares long-horizon predictions. While base predictions become oversmoothed and lose fine-scale vortices, Base+ReAl better preserves these structures throughout the rollout. Additional results are provided in Appendix~\ref{sec:appendix_realPDEBench}.

\begin{figure*}[t]
\centering

\begin{minipage}[t]{0.62\linewidth}
\vspace{0pt}
\centering
\includegraphics[width=\linewidth]{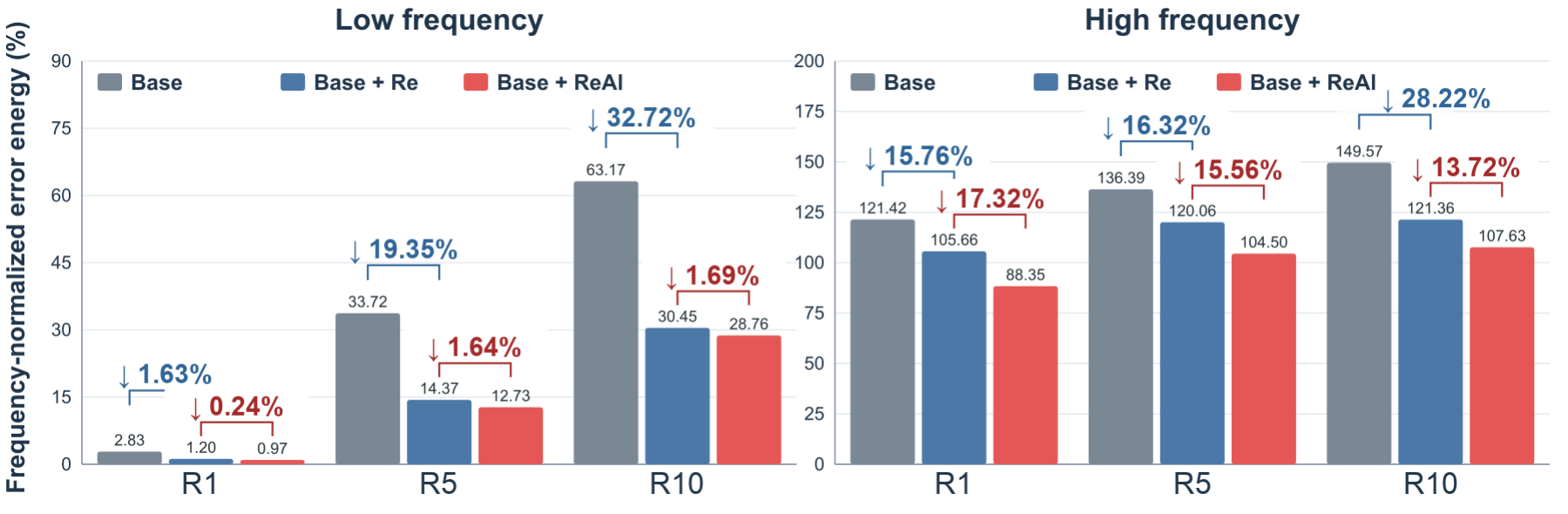}

{\captionsetup{skip=3pt}
\captionof{figure}{Comparison of frequency-normalized error energy across rollout horizons. Lower is better.}
\label{fig:frequency_error_energy}}
\end{minipage}
\hfill
\begin{minipage}[t]{0.34\linewidth}
\vspace{0pt}
\centering
\scriptsize
\setlength{\tabcolsep}{3pt}
\renewcommand{\arraystretch}{1.08}
{\captionsetup{skip=3pt}
\captionof{table}{Numerical frequency-normalized error energy. Lower is better.}
\label{tab:frequency_energy}}

\resizebox{\linewidth}{!}{%
\begin{tabular}{@{}llccc@{}}
\toprule
\textbf{Frequency} & \textbf{Method} & \textbf{R1} & \textbf{R5} & \textbf{R10} \\
\midrule
\multirow{3}{*}{Low}
& Base        & 2.83 & 33.72 & 63.17 \\
& Base + Re   & 1.20 & 14.37 & 30.45 \\
& Base + ReAl & \textbf{0.97} & \textbf{12.73} & \textbf{28.76} \\
\midrule
\multirow{3}{*}{High}
& Base        & 121.42 & 136.39 & 149.57 \\
& Base + Re   & 105.66 & 120.06 & 121.36 \\
& Base + ReAl & \textbf{88.35} & \textbf{104.50} & \textbf{107.63} \\
\bottomrule
\end{tabular}%
}
\end{minipage}
\end{figure*}

\textbf{Frequency Information Recovery.}
To examine how each module corrects autoregressive trajectories, we decompose the prediction error in the radial frequency domain and evaluate the lowest \(20\%\) (\(f<20\)) and highest \(20\%\) (\(f\geq80\)) frequency regions. Here, R1, R5, and R10 denote the first, fifth, and tenth autoregressive rollout blocks, corresponding to rollout horizons of 10, 50, and 100 future frames, respectively. For frequency region \(b\), we use the ground-truth-normalized error energy \(\frac{E_{\mathrm{err},b}}{E_{\mathrm{GT},b}}\times100\%\), where lower values indicate better frequency information recovery. As shown in Fig.~\ref{fig:frequency_error_energy} and Table~\ref{tab:frequency_energy}, the Base+Re correction range expands with the rollout. At R1, it reduces high- and low-frequency errors by \(15.76\%\) and \(1.63\%\), respectively. At R10, these improvements reach \(28.22\%\) and \(32.72\%\), indicating that Base+Re gradually extends from fine-scale to multiscale trajectory correction. In contrast, Base+ReAl consistently targets high-frequency errors: its additional improvements at R1, R5, and R10 are \(17.32\%\), \(15.56\%\), and \(13.72\%\), while the low-frequency gains remain only \(0.24\%\), \(1.64\%\), and \(1.69\%\). Thus, ReAl continuously restores fine-scale information lost during autoregressive prediction.



\newlength{\freqpanelheight}
\setlength{\freqpanelheight}{4.7cm}




\definecolor{basegray}{RGB}{245,245,245}

\subsection{Ablation Study}

\begin{table*}[t]
\centering
\scriptsize
\setlength{\tabcolsep}{2.6pt}
\renewcommand{\arraystretch}{1.15}

\caption{
Reward ablation results on Controlled Cylinder.
$h_r$ denotes the rollout block used to compute the training reward. The best results are shown in \textbf{bold}, and the second-best results are \underline{underlined}.
}
\label{tab:reward_ablation_full_test}

\resizebox{\textwidth}{!}{
\begin{tabular}{llc|ccccc|ccccc|ccccc}
\toprule

\multirow{2}{*}{\textbf{Method}}
& \multirow{2}{*}{\textbf{Reward}}
& \multirow{2}{*}{$\boldsymbol{h_r}$}
& \multicolumn{5}{c|}{\textbf{Evaluation R1}}
& \multicolumn{5}{c|}{\textbf{Evaluation R5}}
& \multicolumn{5}{c}{\textbf{Evaluation R10}}
\\

\cmidrule(lr){4-8}
\cmidrule(lr){9-13}
\cmidrule(lr){14-18}

&&&
\textbf{RMSE} $\downarrow$ &
\textbf{Rel.\ $L_2$} $\downarrow$ &
\textbf{fRMSE} $\downarrow$ &
\textbf{FE} $\downarrow$ &
\textbf{KE} $\downarrow$ &

\textbf{RMSE} $\downarrow$ &
\textbf{Rel.\ $L_2$} $\downarrow$ &
\textbf{fRMSE} $\downarrow$ &
\textbf{FE} $\downarrow$ &
\textbf{KE} $\downarrow$ &

\textbf{RMSE} $\downarrow$ &
\textbf{Rel.\ $L_2$} $\downarrow$ &
\textbf{fRMSE} $\downarrow$ &
\textbf{FE} $\downarrow$ &
\textbf{KE} $\downarrow$
\\

\midrule

\rowcolor{basegray}
Base
& -- & --
& 0.0100 & 0.0750 & 0.0012 & 21.5557 & 3.77e-5
& 0.0227 & 0.1771 & 0.0040 & 46.0869 & 5.57e-5
& 0.0320 & 0.2603 & 0.0065 & 64.3384 & 6.28e-5
\\

+ Re
& -- & --
& 0.0093 & 0.0673 & 0.0011 & 26.3517 & 3.39e-5
& 0.0243 & 0.1830 & 0.0043 & 49.6342 & 5.63e-5
& 0.0331 & 0.2492 & 0.0067 & 65.5248 & 6.33e-5
\\

+ ReAl
& RMSE & 1
& \textbf{0.0082}
& \textbf{0.0602}
& \textbf{0.0010}
& \textbf{20.0997}
& \textbf{3.12e-5}
& 0.0223 & 0.1694 & 0.0039
& \textbf{35.1164}
& 5.49e-5
& 0.0314 & 0.2358 & 0.0063 & 59.6282 & 6.20e-5
\\

+ ReAl
& RMSE & 5
& \underline{0.0084} & 0.0619 & \underline{0.0010} & 20.5995 & 3.22e-5
& \textbf{0.0200}
& \textbf{0.1560}
& \textbf{0.0032}
& 35.4562
& \underline{5.38e-5}
& \underline{0.0273}
& \underline{0.2070}
& \underline{0.0054}
& \textbf{55.0176}
& \underline{5.91e-5}
\\

+ ReAl
& RMSE & 10
& 0.0087 & 0.0637 & 0.0011 & \underline{20.1948} & 3.31e-5
& 0.0210 & 0.1605 & 0.0037 & 35.7731 & 5.45e-5
& \textbf{0.0269}
& \textbf{0.2052}
& \textbf{0.0053}
& \underline{55.2919}
& 5.91e-5
\\

\midrule

+ ReAl
& KE & 5
& 0.0084 & \underline{0.0618} & 0.0010 & 20.7217 & \underline{3.21e-5}
& \underline{0.0208}
& \underline{0.1585}
& \underline{0.0036}
& 35.4098
& \textbf{5.37e-5}
& 0.0273 & 0.2072 & 0.0054 & 55.3497
& \textbf{5.90e-5}
\\

+ ReAl
& FE & 5
& 0.0085 & 0.0624 & 0.0010 & 20.3108 & 3.24e-5
& 0.0210 & 0.1602 & 0.0037
& \underline{35.2623}
& 5.39e-5
& 0.0275 & 0.2095 & 0.0054 & 56.0622 & 5.92e-5
\\

\bottomrule
\end{tabular}
}
\end{table*}


\textbf{Ablation of Reward Horizon and Metric.}
We investigate how the rollout block used to compute the training reward, \(h_r\in\{1,5,10\}\), and the choice of reward metric affect rollout performance. A clear horizon-matching effect emerges: \(h_r=1\), \(5\), and \(10\) achieve the best RMSE at R1, R5, and R10, respectively, while remaining competitive elsewhere. Longer reward horizons improve recursive stability. RMSE provides the most consistent gains in RMSE, Rel. \(L_2\), and fRMSE, while KE yields the lowest kinetic-energy error. FE optimization does not consistently minimize FE, indicating that physics-specific rewards need not improve their corresponding metrics. We therefore use RMSE with \(h_r=5\) by default, which best balances short- and long-horizon performance.

\section{Conclusion}
In this paper, we propose \textbf{Re}fine and \textbf{Al}ign (\textbf{ReAl}), a model-agnostic post-training framework for long-horizon PDE forecasting. \textbf{ReAl} retains an MSE-trained base model to capture stable dominant dynamics and uses a lightweight conditional Flow Refiner to generatively correct prediction residuals, recovering fine-scale high-frequency structures that are often lost under MSE training. We further introduce trajectory-alignment post-training, which optimizes the Refiner using trajectory-level rewards from multiple stochastic autoregressive rollouts to align local corrections with long-term dynamics and suppress accumulated sampling perturbations. We systematically evaluate \textbf{ReAl} on RealPDEBench using UNet, FNO, DPOT-S, and Transolver. \textbf{ReAl} reduces RMSE by \(33.9\%\) on average for $1$-st step predictions and \(45.4\%\) by $20$-th step, which effectively mitigates error accumulation, trajectory divergence, and numerical dissipation over long autoregressive rollouts. Frequency analysis shows that the Refiner gradually extends to multiscale correction, while \textbf{ReAl} mainly provides additional improvements in high-frequency errors. These results demonstrate that \textbf{ReAl} serves as a general, lightweight, and plug-and-play post-training module that improves fine-scale detail recovery and long-horizon prediction stability for diverse PDE surrogates under real-world data distributions and simulation-to-reality shifts.

\newpage
\section*{AI usage statement}
In this work, we used generative AI tools to assist with language
polishing, translation, dataset cleaning and reformatting, qualitative
and thematic data analysis, and the interpretation of experimental
results. We did not use generative AI to generate synthetic datasets,
develop theoretical models or conceptual frameworks, formulate
mathematical claims, provide key elements of or write mathematical
proofs, propose or refine hypotheses, design or provide feedback on
research methodologies or experiments, or implement the proposed
methods. We additionally used generative AI for brainstorming,
information retrieval, improving the readability of the manuscript, and identifying potentially relevant literature. All AI-assisted text and translations were reviewed by the authors, references were verified against their original sources, and data processing, analyses, and interpretations were independently checked against the source data,
implementation, and experimental results. We take full responsibility
for the final content of this work, including all text, claims, and
artifacts produced with the assistance of generative AI.

\newpage

\bibliography{iclr2027_conference}
\bibliographystyle{iclr2027_conference}

\appendix
\newpage

\section{Evaluation Metrics}
\label{sec:appendix_metric}
Let $\widehat{\boldsymbol{Y}}_n,\boldsymbol{Y}_n
\in\mathbb{R}^{C\times H\times W}$ denote the predicted and ground-truth
physical fields at the $n$-th autoregressive rollout step, respectively,
where $C$ is the number of physical variables, and $H$ and $W$ denote
the spatial height and width. All evaluation metrics are computed in the
original physical space. Specifically, we invert the channel-wise
Gaussian normalization as:
\begin{equation}
\widehat{Y}_{n,c,h,w}
=
\widehat{Y}^{\mathrm{norm}}_{n,c,h,w}\sigma_c+\mu_c,
\qquad
Y_{n,c,h,w}
=
Y^{\mathrm{norm}}_{n,c,h,w}\sigma_c+\mu_c,
\label{eq:denormalization}
\end{equation}
where $c\in\{1,\ldots,C\}$ indexes the physical variable, while $h$ and
$w$ index the spatial coordinates. Here,
$\widehat{Y}^{\mathrm{norm}}_{n,c,h,w}$ and
$Y^{\mathrm{norm}}_{n,c,h,w}$ are the normalized prediction and target,
respectively, and $\mu_c$ and $\sigma_c$ are the training-set mean and
standard deviation of the $c$-th physical variable.

Prediction Accuracy Metrics.
We use the Root Mean Squared Error (RMSE) and Relative \(L_2\) Error (Rel. \(L_2\)) to evaluate the pointwise accuracy of the predicted physical fields. At the \(n\)-th rollout step, RMSE is defined as:
\begin{equation}
\operatorname{RMSE}_n
=
\sqrt{
\frac{1}{CHW}
\sum_{c=1}^{C}
\sum_{h=1}^{H}
\sum_{w=1}^{W}
\left(
\widehat{Y}_{n,c,h,w}-Y_{n,c,h,w}
\right)^2
}.
\end{equation}

This metric measures the absolute pointwise discrepancy between the prediction and ground truth over all physical variables and spatial locations.
To account for differences in magnitude across physical variables, we further report the channel-averaged Rel. \(L_2\) Error:
\begin{equation}
\operatorname{Rel.} L_{2,n}
=
\frac{1}{C}
\sum_{c=1}^{C}
\frac{
\left\|
\widehat{\boldsymbol{Y}}_{n,c}
-
\boldsymbol{Y}_{n,c}
\right\|_2
}{
\left\|
\boldsymbol{Y}_{n,c}
\right\|_2
}
=
\frac{1}{C}
\sum_{c=1}^{C}
\sqrt{
\frac{
\sum_{h=1}^{H}\sum_{w=1}^{W}
\left(
\widehat{Y}_{n,c,h,w}-Y_{n,c,h,w}
\right)^2
}{
\sum_{h=1}^{H}\sum_{w=1}^{W}
Y_{n,c,h,w}^{2}
}
}.
\end{equation}
Here, \(\widehat{\boldsymbol{Y}}_{n,c}\) and
\(\boldsymbol{Y}_{n,c}\in\mathbb{R}^{H\times W}\) denote the predicted and ground-truth spatial fields of the \(c\)-th physical variable. Rel. \(L_2\) Error is dimensionless and is computed independently for each physical variable before being averaged across all \(C\) variables.

\noindent\textbf{Physics-Oriented Metrics.}
Beyond pointwise prediction errors, we employ Fourier-space RMSE
(fRMSE), Frequency Error (FE), and Kinetic Energy Error (KE) to evaluate
the spectral and dynamical consistency of the predictions. Lower values
indicate better performance for all three metrics.

\paragraph{Fourier-Space RMSE (fRMSE).}
fRMSE evaluates prediction accuracy in the spatial-frequency domain. At
the $n$-th rollout step, a two-dimensional Fourier transform is applied
independently to each physical variable, and the resulting coefficient
errors are grouped according to their radial frequency magnitudes:
\begin{equation}
\operatorname{fRMSE}_n
=
\frac{1}{CK}
\sum_{c=1}^{C}
\sum_{q=0}^{K-1}
\frac{1}{HW}
\left(
\sum_{(i,j)\in\mathcal{S}_q}
\left|
\mathcal{F}_{h,w}
\left(\widehat{\boldsymbol{Y}}_{n,c}\right)_{i,j}
-
\mathcal{F}_{h,w}
\left(\boldsymbol{Y}_{n,c}\right)_{i,j}
\right|^2
\right)^{\frac{1}{2}}.
\label{eq:frmse}
\end{equation}
Here, $\mathcal{F}_{h,w}$ denotes the two-dimensional spatial Fourier
transform, $(i,j)$ indexes the spatial-frequency coefficients,
$\mathcal{S}_q$ denotes the $q$-th radial frequency shell, and $K$ is
the total number of valid shells. For a more fine-grained spectral
analysis, we further partition the radial spectrum into low-, middle-,
and high-frequency bands.

\paragraph{Frequency Error (FE).}
FE measures whether the predicted trajectory preserves the temporal
frequency characteristics of the underlying dynamics. Let
$\widehat{\boldsymbol{Y}},\boldsymbol{Y}
\in\mathbb{R}^{T\times C\times H\times W}$ denote the predicted and
ground-truth trajectories over $T$ consecutive physical time steps. We
first aggregate each state over all physical variables and spatial
locations:
\begin{equation}
\widehat{s}_t
=
\sum_{c=1}^{C}
\sum_{h=1}^{H}
\sum_{w=1}^{W}
\widehat{Y}_{t,c,h,w},
\qquad
s_t
=
\sum_{c=1}^{C}
\sum_{h=1}^{H}
\sum_{w=1}^{W}
Y_{t,c,h,w}.
\label{eq:temporal_signal}
\end{equation}
A one-dimensional Fourier transform is then applied along the physical
time dimension, and FE is defined as the mean absolute difference
between the predicted and ground-truth temporal spectra:
\begin{equation}
\operatorname{FE}
=
\frac{1}{T}
\sum_{\omega=0}^{T-1}
\left|
\mathcal{F}_{t}(\widehat{\boldsymbol{s}})_{\omega}
-
\mathcal{F}_{t}(\boldsymbol{s})_{\omega}
\right|.
\label{eq:frequency_error}
\end{equation}
Here, $\widehat{\boldsymbol{s}}=(\widehat{s}_1,\ldots,\widehat{s}_T)$
and $\boldsymbol{s}=(s_1,\ldots,s_T)$ are the aggregated temporal
signals, $\mathcal{F}_t$ denotes the one-dimensional Fourier transform
along physical time, and $\omega$ is the corresponding temporal-frequency
index.

\paragraph{Kinetic Energy Error (KE).}
KE evaluates whether the predicted velocity field preserves the
time-averaged fluctuating kinetic-energy distribution. Let $u_{t,h,w}$
and $v_{t,h,w}$ denote the two velocity components at physical time
$t$. Their temporal means and the resulting fluctuating kinetic-energy field are defined as:
\begin{equation}
\begin{aligned}
\overline{u}_{h,w}
&=
\frac{1}{T}\sum_{t=1}^{T}u_{t,h,w},
\qquad
\overline{v}_{h,w}
=
\frac{1}{T}\sum_{t=1}^{T}v_{t,h,w},\\
\mathcal{K}_{h,w}(\boldsymbol{Y})
&=
\frac{1}{2T}
\sum_{t=1}^{T}
\left[
\left(u_{t,h,w}-\overline{u}_{h,w}\right)^2
+
\left(v_{t,h,w}-\overline{v}_{h,w}\right)^2
\right].
\end{aligned}
\label{eq:kinetic_energy_field}
\end{equation}
The kinetic energy error is the spatial mean absolute difference between the predicted and ground-truth energy fields:
\begin{equation}
\operatorname{KE}
=
\frac{1}{HW}
\sum_{h=1}^{H}
\sum_{w=1}^{W}
\left|
\mathcal{K}_{h,w}(\widehat{\boldsymbol{Y}})
-
\mathcal{K}_{h,w}(\boldsymbol{Y})
\right|.
\label{eq:kinetic_energy_error}
\end{equation}
Thus, KE measures the error in the time-averaged fluctuating
kinetic-energy field rather than the instantaneous total kinetic energy. A lower KE indicates better preservation of temporal velocity fluctuations and their spatial energy distribution.

\section{Details of architecture}
\label{sec:appendix_architecture}

We evaluate four representative neural operators under a unified direct multi-step forecasting protocol.

\textbf{U-Net~\citep{ronneberger2015u}}. We use a spatiotemporal U-Net with an encoder–decoder architecture and skip connections. The network contains four resolution scales with channel multipliers \([1,2,4,8]\), corresponding to feature widths \(64,128,256,\) and \(512\) for the \(64\times64\) fields used in our experiments. Downsampling captures large-scale flow structures, while skip connections preserve local details and sharp spatial variations during decoding. The 10-frame input and 10-frame output are processed jointly, enabling the model to learn temporal correlations within a prediction block. The model contains 85.98M parameters for three-field datasets; for Controlled Cylinder, where two additional control variables are included in the input, the model has 86.02M parameters.

\textbf{FNO~\citep{li2020fourier}}. We employ a 3D Fourier Neural Operator that alternates spectral convolution layers with pointwise channel mixing. Spectral convolutions operate jointly over the two spatial dimensions and time, allowing the model to capture long-range interactions efficiently in Fourier space. All FNO models use four Fourier layers. For Controlled Cylinder and Foil, we use width 64 and retained Fourier modes \((4,12,16)\), resulting in 50.36M parameters. For Combustion, we retain \((4,16,16)\) modes with width 64, resulting in 67.14M parameters. For FSI, we use width 128 and \((4,16,16)\) modes, resulting in 268.52M parameters.

\textbf{DPOT-S~\citep{pmlr-v235-hao24d}}. DPOT-S is a  discriminative pretrained operator transformer. It tokenizes spatiotemporal fields into \(8\times8\) patches, maps them into a 1024-dimensional latent space, and applies six operator-Transformer layers with eight internal mixing blocks. Global token interactions and spectral/operator mixing enable the model to capture non-local physical dependencies. DPOT-S contains 30.80M parameters.

\textbf{Transolver~\citep{wu2024transolver}}. We use Transolver as a coordinate-aware Transformer baseline. The model represents each point in the spatiotemporal mesh using its \((x,y,t)\) coordinate and associated physical variables, and models their interactions through slice-based physics attention. Our configuration contains six Transformer layers, hidden dimension 256, eight attention heads, 32 slices, and reference resolution parameter \(4\). The model directly maps the complete 10-frame input block to the complete 10-frame output block. It contains approximately 42.59M parameters.

\textbf{Refiner architecture.} The refiner adopts a time-conditioned 2D U-Net that jointly predicts residuals for all ten future frames. For datasets with three physical variables, the residual trajectory is represented by 30 channels, corresponding to ten frames and three variables. The refiner is conditioned on the concatenation of the deterministic prediction of the ten-frames and the observation history of the ten-frames, resulting in 60 conditioning channels. The U-Net has a base width of 64 and consists of a 256-dimensional sinusoidal time embedding, two downsampling stages, two bottleneck residual blocks, two upsampling stages, and encoder--decoder skip connections. Its six time-conditioned residual blocks each employ Group Normalization, SiLU activations, two $3\times3$ convolutions, and a time-dependent affine projection. The resulting refiner contains 5,03M trainable parameters on standard three-variable datasets and 5,05M parameters on Controlled Cylinder, whose observation history includes two additional control channels.

\section{Results on RealPDEBench}
\label{sec:appendix_realPDEBench}

\begin{figure*}[t]
    \centering
    \includegraphics[width=\textwidth]{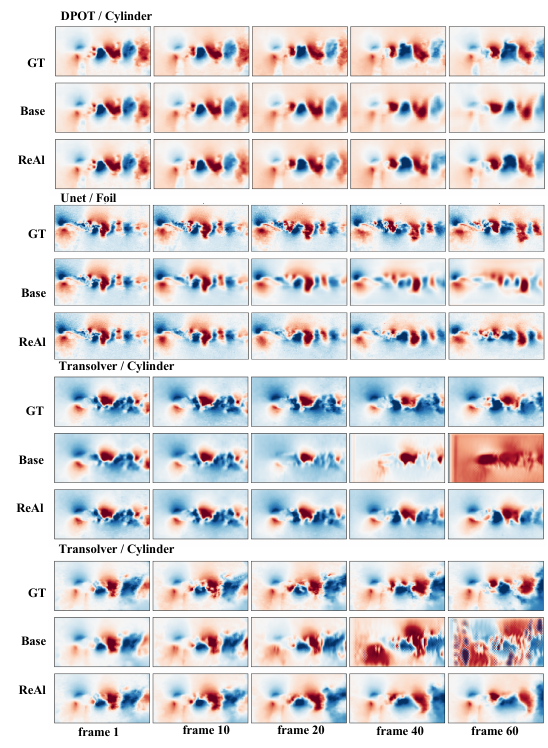}
    \caption{
        More visualizations on the RealPDEBench \textit{Real2Real} setting. Our ReAl can correct the generation trajectory of the Base model.
    }
    \label{fig:supp1}
\end{figure*}

\begin{figure*}[t]
    \centering
    \includegraphics[width=\textwidth]{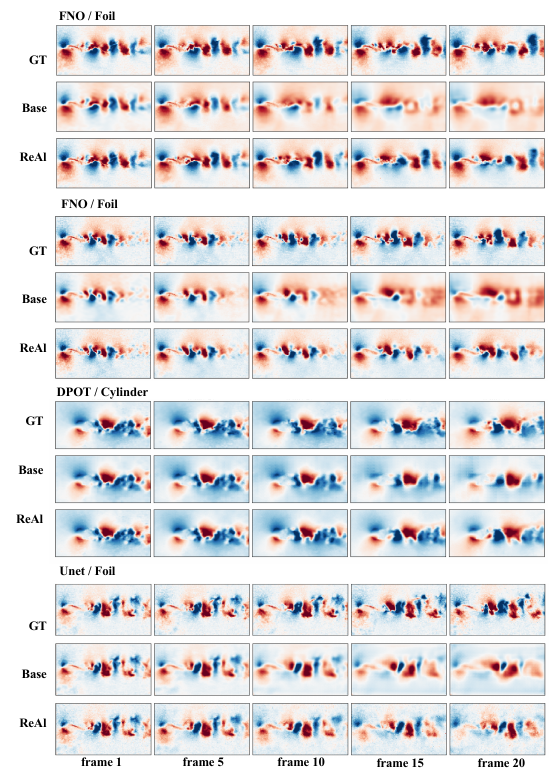}
    \caption{
        Visualizations on the RealPDEBench \textit{Sim2Real} setting. Compared with \textit{Real2Real}, \textit{Sim2Real} involves a substantially larger distribution shift, making autoregressive rollouts more susceptible to error accumulation and long-horizon divergence. We therefore visualize predicted Frames 1, 5, 10, 15, and 20, which cover the informative stage before catastrophic rollout failure. Our ReAl can rectify the generation trajectory of the Base model.
    }
    \label{fig:supp2}
\end{figure*}

Consistent with the quantitative improvements reported in
Table~\ref{tab:main_results}, Table~\ref{tab:app_realpdebench_rmse} further compares RMSE at different
autoregressive rollout horizons. The advantage of ReAl generally becomes more pronounced as the rollout progresses, indicating reduced long-term error accumulation. Table~\ref{tab:app_realpdebench-physics-metrics}
further shows that these improvements extend beyond pointwise prediction accuracy to Rel. $L_2$ error, spectral fidelity, temporal-frequency consistency, and kinetic-energy preservation.  Qualitative comparisons in
Fig.~\ref{fig:supp1} and Fig.~\ref{fig:supp2} show that ReAl effectively
corrects accumulated rollout errors. Under the \textit{Real2Real}
setting, ReAl preserves coherent physical structures over longer
horizons. The \textit{Sim2Real} setting introduces a substantially
larger distribution shift, causing base-model errors to accumulate more rapidly and increasing the risk of trajectory divergence. We therefore visualize Frames 1, 5, 10, 15, and 20, covering the informative rollout period before catastrophic failure. Across both settings, ReAl consistently reduces rollout drift and produces trajectories that remain closer to the ground truth.

\begin{table}[p]
\centering
\caption{RMSE of autoregressive rollout frames. Lower is better; the lowest displayed value within each model--dataset group is shown in bold blue.}
\label{tab:app_realpdebench_rmse}
\scriptsize
\setlength{\tabcolsep}{3.2pt}
\renewcommand{\arraystretch}{1.0}
\begin{adjustbox}{max width=\textwidth,max totalheight=1\textheight,keepaspectratio}
\begin{tabular}{@{}llccccc@{}}
\toprule
\rowcolor{headerblue}
\color{white}Dataset & \color{white}Method & \color{white}R1 & \color{white}R2 & \color{white}R5 & \color{white}R10 & \color{white}R20 \\
\midrule
\rowcolor{groupblue} \multicolumn{7}{l}{\textbf{FNO}} \\
Cylinder & Base & 0.0138 & 0.0192 & 0.0286 & 0.0389 & 0.0572 \\
\rowcolor{rowgray}  & + Re & 0.0104 & 0.0174 & 0.0252 & \best{0.0336} & 0.0492 \\
 & + ReAl & \best{0.0092} & \best{0.0153} & \best{0.0233} & 0.0340 & \best{0.0406} \\
\cmidrule(lr){1-7}
Controlled cylinder & Base & 0.0142 & 0.0181 & 0.0229 & 0.0421 & 0.0584 \\
\rowcolor{rowgray}  & + Re & 0.0120 & 0.0162 & 0.0331 & 0.0407 & 0.0590 \\
 & + ReAl & \best{0.0088} & \best{0.0143} & \best{0.0246} & \best{0.0366} & \best{0.0444} \\
\cmidrule(lr){1-7}
FSI & Base & 0.0091 & 0.0148 & 0.0271 & 0.0687 & 0.9538 \\
\rowcolor{rowgray}  & + Re & 0.0062 & 0.0132 & 0.0266 & 0.0384 & 0.1232 \\
 & + ReAl & \best{0.0058} & \best{0.0124} & \best{0.0254} & \best{0.0370} & \best{0.0510} \\
\cmidrule(lr){1-7}
Foil & Base & 0.0136 & 0.0216 & 0.0407 & 0.0590 & 0.0798 \\
\rowcolor{rowgray}  & + Re & 0.0081 & 0.0131 & 0.0205 & 0.0301 & 0.0455 \\
 & + ReAl & \best{0.0073} & \best{0.0120} & \best{0.0190} & \best{0.0237} & \best{0.0285} \\
\addlinespace[0.35em]
\rowcolor{groupblue} \multicolumn{7}{l}{\textbf{U-Net}} \\
Cylinder & Base & 0.0099 & 0.0164 & 0.0285 & 0.0439 & 0.0691 \\
\rowcolor{rowgray}  & + Re & \best{0.0089} & 0.0147 & 0.0224 & 0.0310 & 0.0413 \\
 & + ReAl & 0.0090 & \best{0.0127} & \best{0.0184} & \best{0.0210} & \best{0.0383} \\
\cmidrule(lr){1-7}
Controlled cylinder & Base & 0.0083 & 0.0135 & 0.0220 & 0.0323 & 0.0463 \\
\rowcolor{rowgray}  & + Re & 0.0077 & 0.0133 & 0.0214 & 0.0307 & 0.0440 \\
 & + ReAl & \best{0.0067} & \best{0.0103} & \best{0.0154} & \best{0.0257} & \best{0.0400} \\
\cmidrule(lr){1-7}
FSI & Base & 0.0059 & 0.0117 & 0.0252 & 0.0424 & 0.0703 \\
\rowcolor{rowgray}  & + Re & 0.0054 & 0.0115 & 0.0248 & 0.0405 & 0.0643 \\
 & + ReAl & \best{0.0052} & \best{0.0085} & \best{0.0208} & \best{0.0325} & \best{0.0603} \\
\cmidrule(lr){1-7}
Foil & Base & 0.0083 & 0.0121 & 0.0225 & 0.0279 & 0.0278 \\
\rowcolor{rowgray}  & + Re & 0.0071 & 0.0115 & 0.0221 & 0.0281 & 0.0303 \\
 & + ReAl & \best{0.0067} & \best{0.0105} & \best{0.0191} & \best{0.0241} & \best{0.0253} \\
\addlinespace[0.35em]
\rowcolor{groupblue} \multicolumn{7}{l}{\textbf{DPOT}} \\
Cylinder & Base & 0.0115 & 0.0181 & 0.0302 & 0.0428 & 0.0679 \\
\rowcolor{rowgray}  & + Re & 0.0105 & 0.0175 & 0.0251 & 0.0331 & 0.0574 \\
 & + ReAl & \best{0.0092} & \best{0.0154} & \best{0.0229} & \best{0.0327} & \best{0.0400} \\
\cmidrule(lr){1-7}
Controlled cylinder & Base & 0.0115 & 0.0191 & 0.0365 & 0.0531 & 0.0643 \\
\rowcolor{rowgray}  & + Re & 0.0091 & 0.0155 & 0.0239 & 0.0351 & 0.0510 \\
 & + ReAl & \best{0.0081} & \best{0.0137} & \best{0.0218} & \best{0.0299} & \best{0.0404} \\
\cmidrule(lr){1-7}
FSI & Base & 0.0106 & 0.0167 & 0.0351 & 0.0544 & 0.0738 \\
\rowcolor{rowgray}  & + Re & 0.0060 & 0.0126 & 0.0254 & 0.0364 & 0.0494 \\
 & + ReAl & \best{0.0056} & \best{0.0118} & \best{0.0237} & \best{0.0344} & \best{0.0399} \\
\cmidrule(lr){1-7}
Foil & Base & 0.0144 & 0.0174 & 0.0284 & 0.1003 & 19.5792 \\
\rowcolor{rowgray}  & + Re & 0.0081 & 0.0130 & 0.0239 & 0.0695 & 0.3616 \\
 & + ReAl & \best{0.0074} & \best{0.0119} & \best{0.0220} & \best{0.0640} & \best{0.0803} \\
\addlinespace[0.35em]
\rowcolor{groupblue} \multicolumn{7}{l}{\textbf{Transolver}} \\
Cylinder & Base & 0.0154 & 0.0248 & 0.0535 & 0.0869 & 0.1101 \\
\rowcolor{rowgray}  & + Re & 0.0104 & 0.0173 & 0.0252 & 0.0441 & 0.0693 \\
 & + ReAl & \best{0.0092} & \best{0.0153} & \best{0.0231} & \best{0.0328} & \best{0.0404} \\
\cmidrule(lr){1-7}
Controlled cylinder & Base & 0.0137 & 0.0232 & 0.0483 & 0.0711 & 0.0921 \\
\rowcolor{rowgray}  & + Re & 0.0094 & 0.0165 & 0.0268 & 0.0518 & 0.0751 \\
 & + ReAl & \best{0.0083} & \best{0.0146} & \best{0.0249} & \best{0.0400} & \best{0.0638} \\
\cmidrule(lr){1-7}
FSI & Base & 0.0119 & 0.0200 & 0.0360 & 0.0570 & 0.1284 \\
\rowcolor{rowgray}  & + Re & 0.0060 & 0.0127 & 0.0260 & 0.0430 & 0.0586 \\
 & + ReAl & \best{0.0056} & \best{0.0118} & \best{0.0243} & \best{0.0368} & \best{0.0554} \\
\cmidrule(lr){1-7}
Foil & Base & 0.0239 & 0.0400 & 0.0849 & 0.0950 & 0.0969 \\
\rowcolor{rowgray}  & + Re & 0.0109 & \best{0.0226} & 0.0323 & 0.0391 & 0.0455 \\
 & + ReAl & \best{0.0100} & 0.0240 & \best{0.0316} & \best{0.0369} & \best{0.0407} \\
\bottomrule
\end{tabular}
\end{adjustbox}
\end{table}

\begin{table}[p]
\centering
\caption{Additional quantitative results across autoregressive rollout steps in the \textit{Real2Real} setting. Each rollout step aggregates 10 frames. Lower is better; entries highlighted in the supplied table are shown in bold blue.}
\label{tab:app_realpdebench-physics-metrics}
\fontsize{6.8}{7.4}\selectfont
\setlength{\tabcolsep}{1.8pt}
\renewcommand{\arraystretch}{1.35}
\begin{adjustbox}{width=\textwidth,max totalheight=0.84\textheight,keepaspectratio,center}
\begin{tabular}{@{}llcccccccccccc@{}}
\toprule
\rowcolor{headerblue}
\color{white}Dataset & \color{white}Method & \multicolumn{4}{c}{\color{white}R1} & \multicolumn{4}{c}{\color{white}R5} & \multicolumn{4}{c}{\color{white}R10} \\
\cmidrule(lr){3-6} \cmidrule(lr){7-10} \cmidrule(lr){11-14}
\rowcolor{headerblue}
 & & \color{white}Rel $L_2$ & \color{white}fRMSE & \color{white}FE & \color{white}KE & \color{white}Rel $L_2$ & \color{white}fRMSE & \color{white}FE & \color{white}KE & \color{white}Rel $L_2$ & \color{white}fRMSE & \color{white}FE & \color{white}KE \\
\midrule
\rowcolor{groupblue} \multicolumn{14}{l}{\textbf{FNO}} \\
Cylinder & Base & 0.074 & 2.02e-3 & 52.61 & 6.65e-5 & 0.161 & 5.77e-3 & 154.37 & 1.13e-4 & 0.220 & 8.45e-3 & 219.35 & 1.55e-4 \\
\rowcolor{rowgray}  & + Re & 0.053 & 1.88e-3 & 42.55 & 6.00e-5 & 0.143 & 4.10e-3 & 100.95 & 9.37e-5 & 0.197 & 7.05e-3 & 87.12 & 1.30e-4 \\
 & + ReAl & \best{0.047} & \best{1.11e-3} & \best{28.92} & \best{4.90e-5} & \best{0.123} & \best{3.96e-3} & \best{55.08} & \best{8.84e-5} & \best{0.175} & \best{6.46e-3} & \best{64.18} & \best{1.07e-4} \\
\cmidrule(lr){1-14}
Controlled cylinder & Base & 0.073 & 1.17e-3 & 21.63 & 3.69e-5 & 0.173 & 3.96e-3 & 45.72 & 5.43e-5 & 0.251 & 6.42e-3 & 146.69 & 6.14e-5 \\
\rowcolor{rowgray}  & + Re & 0.066 & 1.06e-3 & 22.84 & 3.31e-5 & 0.178 & 4.21e-3 & 46.84 & 5.49e-5 & 0.242 & 6.55e-3 & 65.37 & 6.21e-5 \\
 & + ReAl & \best{0.058} & \best{9.32e-4} & \best{20.90} & \best{3.02e-5} & \best{0.165} & \best{3.84e-3} & \best{44.41} & \best{5.33e-5} & \best{0.229} & \best{6.14e-3} & \best{55.47} & \best{6.01e-5} \\
\cmidrule(lr){1-14}
FSI & Base & 0.066 & 7.50e-4 & \best{7.12} & 1.85e-5 & 0.185 & 3.64e-3 & 36.53 & 3.32e-5 & 0.379 & 1.22e-2 & 299.36 & 2.27e-3 \\
\rowcolor{rowgray}  & + Re & 0.054 & 4.25e-4 & 10.58 & 1.50e-5 & 0.181 & 3.28e-3 & \best{21.49} & 2.31e-5 & 0.303 & 6.09e-3 & 56.44 & 2.51e-5 \\
 & + ReAl & \best{0.041} & \best{4.11e-4} & 10.67 & \best{1.41e-5} & \best{0.174} & \best{3.15e-3} & 21.57 & \best{2.23e-5} & \best{0.254} & \best{5.89e-3} & \best{31.44} & \best{2.43e-5} \\
\cmidrule(lr){1-14}
Foil & Base & 0.029 & 2.16e-3 & 95.47 & 5.17e-5 & 0.087 & 7.42e-3 & 321.41 & 9.04e-5 & 0.126 & 1.09e-2 & 466.20 & 9.94e-5 \\
\rowcolor{rowgray}  & + Re & 0.015 & 9.81e-4 & 30.57 & 4.74e-5 & 0.053 & 3.28e-3 & 60.71 & 7.97e-5 & 0.070 & 5.26e-3 & 80.15 & 9.31e-5 \\
 & + ReAl & 0.014 & \best{5.55e-4} & \best{24.15} & \best{3.39e-5} & \best{0.031} & \best{2.18e-3} & \best{45.16} & \best{7.16e-5} & \best{0.038} & \best{3.15e-3} & \best{60.07} & \best{8.40e-5} \\
\addlinespace[0.25em]
\rowcolor{groupblue} \multicolumn{14}{l}{\textbf{U-Net}} \\
Cylinder & Base & 0.051 & 1.28e-3 & 27.48 & 5.14e-5 & 0.153 & 5.55e-3 & 94.11 & 1.00e-4 & 0.234 & 9.57e-3 & 172.41 & 1.26e-4 \\
\rowcolor{rowgray}  & + Re & 0.040 & 1.13e-3 & 25.84 & 5.00e-5 & 0.139 & 4.00e-3 & 65.91 & 8.96e-5 & 0.207 & 6.86e-3 & 100.83 & 1.13e-4 \\
 & + ReAl & \best{0.045} & \best{1.06e-3} & \best{19.84} & \best{4.77e-5} & \best{0.119} & \best{3.62e-3} & \best{40.91} & \best{8.66e-5} & \best{0.167} & \best{5.56e-3} & \best{52.83} & \best{1.03e-4} \\
\cmidrule(lr){1-14}
Controlled cylinder & Base & 0.057 & 9.06e-4 & 17.12 & 3.02e-5 & 0.159 & 3.87e-3 & 35.67 & 5.28e-5 & 0.235 & 6.66e-3 & 50.90 & 6.20e-5 \\
\rowcolor{rowgray}  & + Re & 0.054 & 9.00e-4 & 16.51 & 2.91e-5 & 0.152 & 3.74e-3 & 34.34 & 5.26e-5 & 0.228 & 6.43e-3 & 49.22 & 6.18e-5 \\
 & + ReAl & \best{0.051} & \best{8.70e-4} & \best{16.11} & \best{2.81e-5} & \best{0.152} & \best{3.70e-3} & \best{33.94} & \best{5.26e-5} & \best{0.218} & \best{6.33e-3} & \best{48.22} & \best{6.16e-5} \\
\cmidrule(lr){1-14}
FSI & Base & 0.042 & 3.34e-4 & \best{3.43} & 1.39e-5 & 0.173 & 3.00e-3 & 13.87 & 2.24e-5 & 0.293 & 6.77e-3 & 29.24 & 2.57e-5 \\
\rowcolor{rowgray}  & + Re & 0.039 & 3.29e-4 & 3.53 & 1.45e-5 & \best{0.170} & 2.93e-3 & 13.56 & \best{2.23e-5} & 0.279 & 6.53e-3 & 27.73 & 2.55e-5 \\
 & + ReAl & \best{0.036} & \best{3.21e-4} & 3.49 & \best{1.35e-5} & \best{0.170} & \best{2.93e-3} & \best{13.34} & \best{2.23e-5} & \best{0.201} & \best{6.33e-3} & \best{26.73} & \best{2.54e-5} \\
\cmidrule(lr){1-14}
Foil & Base & \best{0.014} & 6.73e-4 & 12.94 & 3.53e-5 & 0.036 & 2.33e-3 & 22.92 & 8.71e-5 & \best{0.044} & 3.44e-3 & 33.74 & 1.07e-4 \\
\rowcolor{rowgray}  & + Re & \best{0.012} & 5.99e-4 & 12.46 & 3.32e-5 & \best{0.034} & 2.33e-3 & 22.54 & 8.67e-5 & \best{0.043} & 3.53e-3 & 33.88 & 1.06e-4 \\
 & + ReAl & \best{0.012} & \best{5.51e-4} & \best{12.06} & \best{3.15e-5} & \best{0.034} & \best{2.23e-3} & \best{22.44} & \best{8.43e-5} & \best{0.040} & \best{3.35e-3} & \best{33.26} & \best{1.06e-4} \\
\addlinespace[0.25em]
\rowcolor{groupblue} \multicolumn{14}{l}{\textbf{DPOT}} \\
Cylinder & Base & 0.061 & 1.25e-3 & 28.83 & 6.64e-5 & 0.164 & 5.40e-3 & 66.68 & 9.76e-5 & 0.239 & 8.59e-3 & 114.26 & 1.11e-4 \\
\rowcolor{rowgray}  & + Re & 0.054 & 1.27e-3 & 29.56 & 5.54e-5 & 0.133 & 4.11e-3 & 56.62 & 9.43e-5 & 0.176 & \best{6.01e-3} & 79.60 & 1.12e-4 \\
 & + ReAl & \best{0.048} & \best{1.12e-3} & \best{28.26} & \best{4.95e-5} & \best{0.123} & \best{3.84e-3} & \best{53.99} & \best{8.82e-5} & \best{0.172} & 6.04e-3 & \best{64.48} & \best{1.09e-4} \\
\cmidrule(lr){1-14}
Controlled cylinder & Base & 0.083 & 1.21e-3 & 33.34 & 4.98e-5 & 0.253 & 7.19e-3 & 65.26 & 6.41e-5 & 0.368 & 1.22e-2 & 111.17 & 6.62e-5 \\
\rowcolor{rowgray}  & + Re & 0.062 & 9.94e-4 & 25.05 & 3.86e-5 & 0.229 & 5.04e-3 & 44.68 & 5.42e-5 & 0.342 & 8.91e-3 & \best{62.47} & 6.26e-5 \\
 & + ReAl & \best{0.055} & \best{8.84e-4} & \best{20.25} & \best{2.90e-5} & \best{0.154} & \best{3.73e-3} & \best{43.68} & \best{5.25e-5} & \best{0.226} & \best{6.63e-3} & 66.28 & \best{6.06e-5} \\
\cmidrule(lr){1-14}
FSI & Base & 0.078 & 4.87e-4 & \best{8.29} & 2.10e-5 & 0.236 & 4.95e-3 & 26.01 & 2.50e-5 & 0.367 & 9.57e-3 & 46.48 & 2.53e-5 \\
\rowcolor{rowgray}  & + Re & 0.043 & 3.95e-4 & 9.05 & 1.46e-5 & 0.173 & 3.06e-3 & 21.13 & 2.26e-5 & 0.247 & 5.61e-3 & 30.59 & 2.45e-5 \\
 & + ReAl & \best{0.041} & \best{3.75e-4} & 9.07 & \best{1.37e-5} & \best{0.162} & \best{2.85e-3} & \best{20.60} & \best{2.19e-5} & \best{0.234} & \best{5.31e-3} & \best{29.08} & \best{2.40e-5} \\
\cmidrule(lr){1-14}
Foil & Base & 0.024 & 1.15e-3 & 45.23 & 8.14e-5 & 0.047 & 4.34e-3 & 120.24 & 1.09e-4 & 0.087 & 1.22e-2 & 424.58 & 9.59e-4 \\
\rowcolor{rowgray}  & + Re & 0.015 & 6.43e-4 & 26.53 & 3.75e-5 & 0.036 & 3.00e-3 & 80.07 & 1.10e-4 & 0.067 & 7.97e-3 & 351.05 & 5.78e-4 \\
 & + ReAl & \best{0.015} & \best{6.15e-4} & \best{26.11} & \best{3.42e-5} & \best{0.033} & \best{2.78e-3} & \best{73.23} & \best{9.37e-5} & \best{0.063} & \best{7.39e-3} & \best{231.77} & \best{2.46e-4} \\
\addlinespace[0.25em]
\rowcolor{groupblue} \multicolumn{14}{l}{\textbf{Transolver}} \\
Cylinder & Base & 0.079 & 2.02e-3 & 45.19 & 7.41e-5 & 0.291 & 1.06e-2 & 381.60 & 5.78e-4 & 0.475 & 1.93e-2 & 986.86 & 2.80e-3 \\
\rowcolor{rowgray}  & + Re & 0.053 & 1.26e-3 & 30.37 & 5.45e-5 & 0.173 & 4.25e-3 & 169.53 & 1.28e-4 & 0.279 & 6.49e-3 & 187.00 & 3.09e-4 \\
 & + ReAl & \best{0.047} & \best{1.11e-3} & \best{28.29} & \best{4.89e-5} & \best{0.122} & \best{3.97e-3} & \best{62.84} & \best{8.72e-5} & \best{0.172} & \best{6.36e-3} & \best{75.65} & \best{1.05e-4} \\
\cmidrule(lr){1-14}
Controlled cylinder & Base & 0.096 & 1.86e-3 & 50.65 & 4.63e-5 & 0.339 & 1.13e-2 & 422.53 & 3.20e-4 & 0.542 & 1.93e-2 & 817.56 & 6.55e-4 \\
\rowcolor{rowgray}  & + Re & 0.074 & 1.10e-3 & 28.71 & 3.26e-5 & 0.195 & 5.46e-3 & 92.06 & 5.64e-5 & 0.300 & 9.39e-3 & 157.28 & 6.51e-5 \\
 & + ReAl & \best{0.057} & \best{9.89e-4} & \best{27.24} & \best{2.97e-5} & \best{0.170} & \best{5.05e-3} & \best{85.94} & \best{5.43e-5} & \best{0.261} & \best{9.09e-3} & \best{140.25} & \best{6.31e-5} \\
\cmidrule(lr){1-14}
FSI & Base & 0.082 & 1.25e-3 & 13.07 & 1.82e-5 & 0.246 & 5.67e-3 & 55.36 & 4.63e-5 & 0.406 & 1.19e-2 & 134.49 & 2.64e-4 \\
\rowcolor{rowgray}  & + Re & 0.063 & 4.13e-4 & 9.58 & 1.46e-5 & 0.196 & 3.45e-3 & 32.82 & 2.32e-5 & 0.294 & 6.93e-3 & 75.85 & 2.56e-5 \\
 & + ReAl & \best{0.041} & \best{3.83e-4} & \best{9.35} & \best{1.37e-5} & \best{0.166} & \best{3.24e-3} & \best{31.66} & \best{2.23e-5} & \best{0.251} & \best{6.58e-3} & \best{54.77} & \best{2.46e-5} \\
\cmidrule(lr){1-14}
Foil & Base & 0.039 & 3.91e-3 & 75.26 & 1.04e-4 & 0.115 & 1.89e-2 & 422.58 & 3.12e-4 & 0.128 & 1.93e-2 & 427.76 & 3.06e-4 \\
\rowcolor{rowgray}  & + Re & 0.029 & 1.33e-3 & 41.19 & 5.66e-5 & 0.083 & 4.69e-3 & \best{83.06} & 8.64e-5 & 0.070 & \best{5.45e-3} & 193.00 & 1.19e-4 \\
 & + ReAl & \best{0.018} & \best{1.26e-3} & \best{39.04} & \best{5.22e-5} & \best{0.041} & \best{4.32e-3} & 84.29 & \best{8.15e-5} & \best{0.049} & 5.52e-3 & \best{96.37} & \best{8.16e-5} \\
\bottomrule
\end{tabular}
\end{adjustbox}
\end{table}

\end{document}